\documentclass{article}
\pdfoutput=1

\usepackage{arxiv}

\usepackage{natbib}
 \setcitestyle{aysep={}} 
 \renewcommand{\cite}{\citep}    

\usepackage[utf8]{inputenc} 
\usepackage[T1]{fontenc}    
\usepackage{hyperref}       
\usepackage{url}            
\usepackage{booktabs}       
\usepackage{amsfonts}       
\usepackage{nicefrac}       
\usepackage{microtype}      
\usepackage{lipsum}

\usepackage{synttree}
\usepackage{tikz-dependency}
\usepackage{amsmath}
\usepackage{wasysym}
\usepackage{vcell}

\usepackage{hyperref}
\usepackage{xcolor}
\definecolor{darkblue}{rgb}{0, 0, 0.5}
\hypersetup{colorlinks=true,citecolor=darkblue, linkcolor=darkblue, urlcolor=darkblue}

\title{Measuring Grammatical Diversity from Small Corpora: Derivational Entropy Rates, Mean Length of Utterances, and Annotation Invariance}

\author{Ferm\'{\i}n Moscoso del Prado Mart\'{\i}n\\
Department of Computer Science and Technology \& Jesus College \\
University of Cambridge, UK \\
\texttt{fm611@cst.cam.ac.uk.}}

\begin{document}
\maketitle

\begin{abstract}
In many fields, such as language acquisition, neuropsychology of language, the study of aging, and historical linguistics, corpora are used for estimating the diversity of grammatical structures that are produced during a period by an individual, community, or type of speakers. In these cases, treebanks are taken as representative samples of the syntactic structures that might be encountered. Generalizing the potential syntactic diversity from the structures documented in a small corpus requires careful extrapolation whose accuracy is constrained by the limited size of representative sub-corpora. In this article, I demonstrate --theoretically, and empirically-- that a grammar's derivational entropy and the mean length of the utterances (MLU) it generates are fundamentally linked, giving rise to a new measure, the derivational entropy rate. The mean length of utterances becomes the most practical index of syntactic complexity; I demonstrate that MLU is not a mere proxy, but a fundamental measure of syntactic diversity. In combination with the new derivational entropy rate measure, it provides a theory-free assessment of grammatical complexity. The derivational entropy rate indexes the rate at which different grammatical annotation frameworks determine the grammatical complexity of treebanks. I introduce the Smoothed Induced Treebank Entropy (SITE) as a tool for estimating these measures accurately, even from very small treebanks. I conclude by discussing important implications of these results for both NLP and human language processing.
\end{abstract}

\section{Introduction}

Estimating the diversity of the grammatical structures --sometimes referred to as `syntactic complexity'-- produced by an individual or group thereof is important in multiple areas of linguistics and psychology. Frequent applications include --among others-- drawing inferences on the process of language acquisition by children \cite*[e.g.][]{Theakston:etal:2004} or second language learners \cite*[e.g.][]{Norris:Ortega:2009,Jiang:etal:2019}, assessing the severity and progress of neuropsychological disorders \cite*[e.g.][]{Roark:etal:2007,Pakhomov:etal:2011,Rezaii:etal:2022},  investigating the changes in a language along its history \cite[e.g.][]{Moscoso:2014} or along an individual's lifetime \cite*[e.g.][]{Agmon:etal:2024}. Typically, researchers obtain relevant samples of language (i.e. recorded and transcribed conversations, writings, etc.) which are then syntactically annotated, either semi-automatically using parsers, or manually by trained linguists. The syntactic annotations used in these resources are most frequently in the form of context-free derivation trees \cite*[e.g.][]{Marcus:etal:1995} or --more recently-- dependency graphs \cite*[e.g.][]{deMarneffe:etal:2021}.

Generally speaking, the number of distinct syntactic structures that an individual might be able to produce is --according to some \cite[see][for discussion]{Pullum:Scholz:2010}-- not finite, and certainly extremely large. Researchers therefore need to generalize how much actual grammatical knowledge is represented by a collection of observed syntactic structures. Multiple techniques have been developed in the literature to try to measure the amount of syntactic knowledge represented by a corpus. On the one hand, some researchers have measured the syntax `by proxy', using indices that would correlate with the diversity of syntactic structures. The most common of such indices are the {\bf mean length of utterance} in words \cite[{\bf MLU};][]{Nice:1925} or morphemes \cite{Brown:1973}. On the other hand, some have turned to measures of structural properties argued to be costly by specific theories of syntactic processing \cite[e.g.][]{Yngve:1960,Frazier:1985}, or the mean/total number of words intervening between any two syntactic dependents \cite{Lin:1996,Gibson:1998,Liu:2008}. Both of these types of approaches fall short of looking at the actual diversity of syntactic structures. Finally, a third group have developed indices counting the presence/absence or the number of times that certain grammatical phenomena argued to be of importance are encountered \cite[e.g.  subordination, center-embedding, etc.;][]{Scarborough:1990,Agmon:etal:2024}, assigning some specific diversity values to each occurrence of each phenomenon. Notice, however, that choosing some specific syntactic constructions as the keystones of diversity makes such measures difficult to apply in languages that do not use such mechanisms, and indeed, none of the proposed constructions can be taken as `language universals' \cite{Evans:Levinson:2009}. Even disregarding the heavy annotation burden that such approach imposes, relying on the occurrence of specific constructions for measuring diversity will render the results incomparable across languages. In summary, each of these approaches to measuring diversity presents a number of advantages and inconvenients \cite[see][for overviews of different measures]{Cheung:Kemper:1992,Agmon:etal:2024}. All these approaches share an ad hoc nature very specific to certain areas of linguistic research, and even to individual languages. 

In contrast, some researchers have turned to more natural, standard information theorical measures for the diversity of syntactic structures. The entropy \cite{Shannon:1948} of a population is the  measure of diversity most commonly used across a variety disciplines, such as Economics \cite[e.g.][]{Attaran:Zwick:1987}, Ecology \cite[e.g.][]{Gotelli:Chao:2013}, Genetics \cite*[e.g.][]{Poon:etal:2016}, and is also used for measuring other aspects in Linguistics, such as vocabulary richness \cite{Moscoso:2014}, the size of derivational \cite{Moscoso:cognition:2004} or inflectional \cite{Baayen:Moscoso:2005} morphological paradigms, quantifying the degree of syntactic word order variations \cite{Futtrell:etal:2015,Levshina:2019} or language sequence predictability \cite*{Roark:etal:2007,Sy:etal:2023}. In contrast with those applications, where diversity is related to the distribution of a finite --even if perhaps very large-- number of alternatives (e.g. words, species, etc.), syntactic constructions cannot be thought in terms of their raw number: Even extremely simple grammatical structures can give rise to an infinite number of alternatives. Perhaps the most widely used method for representing syntactic structures and their probabilistic properties are {\bf Probabilistic Context-Free Grammars} \cite[{\bf PCFGs};][]{Booth:1969,Booth:Thompson:1973}. Given a PCFG that has been induced from a treebank, it is known that the entropy of the possibly infinite set syntactic structures it generates is finite and can be determined in closed form \cite{Grenander:1967,Grenander:1976,Miller:OSullivan:1992,Chi:1999}. This {\bf derivational entropy} has been shown to account for the varying levels of difficulty that people encounter when comprehending language, correlating with experimental measurements of sentence reading times \cite{Hale:2003,Linzen:Jaeger:2014}, human memory accuracy for sentences \cite{Hale:2006}, the complexity of the syntax used by adults aging \cite{Moscoso:aging:2017}, or across historical periods \cite{Moscoso:2014}, and --within Computational Linguistics proper-- the degree of difficulty one encounters when parsing different languages \cite{Corazza:etal:2013}.

Researchers often need to generalize the observed syntactic structures, not only those that are explicitly documented in the sample, but also those that could eventually be produced by the individual or population being studied. In these cases, the PCFG is not available a~priori, but it is still necessary to draw conclusions beyond the sentences that were directly observed. A powerful tool is to then induce a PCFG from the observed treebank, and then estimate its entropy as a measure of its diversity, and by extension of the diversity of the structures that \emph{could} have been encountered. Such generalization step comes with its own problems. Typically, a PCFG is induced from a treebank using the method of maximum likelihood (ML). This estimation approach is known to be consistent: With increasing corpus sizes, the estimated rule probabilities converge to their true values with probability 1 \cite{Sanchez:Benedi:1997,Chi:Geman:1998}. In turn, the computation of the PCFG entropy from the rule probabilities is exact \cite{Grenander:1967,Grenander:1976,Miller:OSullivan:1992,Chi:1999}, and therefore the entropy estimator based on the induced grammar is the ML estimate of the entropy and must also be consistent, eventually converging to true PCFG entropy value. The problem arises because, even if ultimately converging, ML entropy estimates are known to be very biased \cite{Miller:1955}: ML consistently underestimates true entropy values. The degree of underestimation correlates positively with two factors: (a) the size of the sample on which the estimates were computed, and (b) the number of possible values of the discrete random variable the entropy of whose distribution is being estimated. The second of these problems could not actually get worse in the case of PCFGs; the number of possible alternatives (the syntactic structures) is, in most cases, infinite. This is further aggravated by the Zipfian power-law nature of their distribution, by which some structures will only be observed extremely rarely. One can therefore expect the degree of underestimation to be rather substantial. The presence of this bias has even motivated some researchers to outright avoid using entropy-based measures of syntactic diversity \cite[e.g.][]{Jing:etal:2021}. In many applications this problem is aggravated by the sample size (i.e. the size of the treebank used for entropy estimation) being very small, and often the sample size being itself related to the very properties of the individual or population that one is interested in studying. For instance, when studying child language acquisition, or language changes during aging, the available samples of utterances produced by an individual will be necessarily small. First, it is very costly to collect and annotate such samples, and that requires a certain degree of intrusion into the individuals' lives. Second, some individuals (e.g. younger children, older males) tend to talk less than others. This introduces a systematic relation between a person's age and the sample size available \cite{Spokoyny:2016,Moscoso:aging:2017}. If one wants to draw inferences on the change in syntactic abilities with age, using ML estimates of derivational entropy one will encounter a confound between changes in actual syntactic diversity, and mere changes in sample size (i.e. being more or less talkative). Such systematic relations with the amount of language produced will also arise when one is studying other populations, such as people suffering from neuropsychological disorders. 

In this study, I demonstrate how the mean length of utterances and the derivational entropies inferred from treebanks are fundamentally related through the concept of derivational entropy rate, and this has far reaching implications for measuring syntactic diversity from empirical samples. In what follows, I begin by establishing the notation and outlining some basic concepts in formal language theory and information theory (Section~\ref{sec:notation}). This is followed, in Section~\ref{sec:theory}, by developing the theoretical link between the mean length of utterances and the derivational entropy of a grammar, introducing the concept of derivational entropy rate. Here, I put forward two hypotheses about this relationship and its implications. Section~\ref{sec:SITEsec} introduces a method for derivational entropy estimation from treebanks: the Smoothed Induced Treebank Entropy (SITE). This measure, first used in \citet{Moscoso:aging:2017}, builds on the hybrid approach suggested by \citet{Moscoso:2014}, and combines it with a better method for bias correction in small datasets \cite{Chao:etal:2013}. In several of corpus analyses (on both context-free and dependency treebanks), described in Section~\ref{sec:corpusan}, I demonstrate the accuracy and the limits of SITE for estimating derivational entropies, and I test the theoretical predictions of the relationship between mean length of utterance and derivational entropy.  Finally, I conclude in Section~\ref{sec:discussion} by discussing the value and further implications of these results.

\section{Notation and Basic Concepts}
\label{sec:notation}

\subsection{Probabilistic Contex-Free Grammars}

A PCFG \cite{Booth:1969,Booth:Thompson:1973} is a quadruple $G=(T,N,S,R)$. In this notation, $T=\{a_1,\ldots,a_{\|T\|}\}$ is a finite set of {\bf terminal symbols}, often referred to as the {\bf alphabet} on which the grammar is defined (here the operator $\| \cdot \|$ denotes the cardinality of a set). $N=\{A_1,\ldots,A_{\|N\|}\}$ is a finite set of {\bf non-terminal symbols}, from which $S \in N$ is the {\bf starting symbol}, that is, the non-terminal that serves as the root of all parse trees generated by the grammar. Finally, $R$ is a set of rules $R=\{r_1,\ldots,r_{\|R\|}\}$ of the form $r \equiv p:\, A \to \alpha$, with the left-hand side of the arrow being a non-terminal symbol ($A \in N$), its right-hand side being a string of terminal and non-terminal symbols ($\alpha \in [N \cup T]^*$), and $0 \leq p \leq 1$ denoting a probability value for the rule. Each of these rules states that, in a string of terminals and/or non-terminals, rewriting the non-terminal $A$ with the string $\alpha$ is an operation allowed by the grammar occurring with probability $p$.

For each non-terminal symbol $A_i$, $R_i \subseteq R$ denotes the subset of $R$ consisting of the $\|R_i\|$ rules that have $A_i$ in their left-hand side. For each non-terminal symbol, I assume a rule ordering such that the set of its rules can be denoted as $R_i = \{r_{i,1},\ldots,r_{i,\|R_i\|}\}$. In this way $r_{i,j}$ always refers to the $j$-th rule having the non-terminal $A_i$ as its left-hand side, and $p_{i,j}$ refers to the probability associated to that rule. The probabilities for the rules expanding each non-terminal symbol in a PCFG need to be proper, in the sense that they are normalized such that
\begin{equation}
\sum_{r_{i,j} \in R_{i}} p_{i,j} = 1, \qquad \text{for all } 1 \leq i \leq \|N\|
\end{equation}
The PCFG generates derivation parse trees through recursive application of its production rules, beginning with the starting symbol $S$ until one obtains a sequence composed only of terminal symbols. A PCFG can generate in this way a possibly infinite set of trees $\mathcal{T}[G] = \{t_1, t_2, t_3, \ldots \}$. The probabilities associated with the rules also associate a probability value to each parse tree; this is given by
\begin{equation}
\mathrm{p}_G(t) = \prod_{r_i \in R} p_i^{f(r_i; t)}\label{eq:treeprob}
\end{equation}
where $f(r_i; t)$ denotes the number of applications of rule $r_i$ involved in the construction of the tree $t$. The PCFG itself also needs to be {\bf proper}, in the sense that 
\begin{equation}
\sum_{t \in \mathcal{T}(G)} \mathrm{p}_G(t) = 1 \label{eq:PCFGproper}
\end{equation}
In fact, it is possible to create improper PCFGs, but any PCFG that is induced from a treebank by maximum-likelihood will always be proper \cite{Chi:1999,Chi:Geman:1998,Sanchez:Benedi:1997}. 

\subsection{Maximum-Likelihood Induction of a PCFG from a Treebank\label{sec:MLPCFG}}

\begin{figure*}[th]
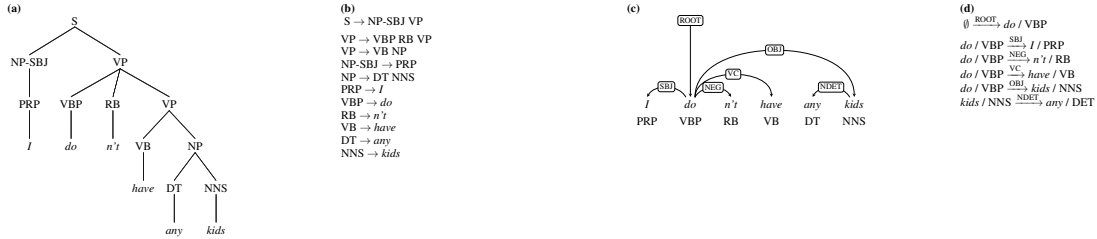

\resizebox{\textwidth}{!}{
\begin{tabular}{cccccc}
\begin{minipage}[t]{.5\textwidth}
{\bf (a)}\\
\vspace{.4\baselineskip}
\synttree[S [NP-SBJ [PRP [{\em I}]]] [VP [VBP [{\em do}]] [RB [{\em n't}]] [VP [VB [{\em have}]] [NP [DT [{\em any}]] [NNS [{\em kids}]]]]]]
\end{minipage}  & \hspace{10mm} &
\begin{minipage}[t]{.5\textwidth}
{\bf (b)}\\
\vspace{.4\baselineskip}
S  $\to$  NP-SBJ  VP  \\
VP  $\to$  VBP  RB  VP \\
VP  $\to$  VB  NP \\
NP-SBJ  $\to$  PRP \\
NP  $\to$  DT  NNS \\
PRP $\to$ \emph{I} \\
VBP $\to$ \emph{do} \\
RB $\to$ \emph{n't} \\
VB $\to$ \emph{have} \\
DT $\to$ \emph{any} \\
NNS $\to$ \emph{kids}
\end{minipage} &
\begin{minipage}[t]{.5\textwidth}
{\bf (c)}\\
\vspace{.4\baselineskip}
\begin{dependency}[arc edge]
\begin{deptext}[column sep=0.5cm, row sep=.5ex]
{\em I} \& {\em do} \& {\em n't} \& {\em have} \& {\em any} \& {\em kids} \\
PRP \& VBP \& RB \& VB \& DT \& NNS \\
\end{deptext}
\depedge{2}{1}{SBJ}
\depedge{2}{3}{NEG}
\depedge[arc angle=88]{2}{4}{VC}
\depedge[arc angle=90]{2}{6}{OBJ}
\depedge{6}{5}{NDET}
\deproot[edge unit distance=4ex]{2}{ROOT}
\end{dependency}
\end{minipage} & \hspace{10 mm}
&
\begin{minipage}[t]{.5\textwidth} 
{\bf (d)}\\
\vspace{.4\baselineskip}
$\emptyset$ $\xrightarrow{\textrm{ROOT}}$ \emph{do} / VBP \\
\emph{do} / VBP $\xrightarrow{\textrm{SBJ}}$  \emph{I} / PRP  \\
\emph{do} / VBP  $\xrightarrow{\textrm{NEG}}$  \emph{n't} / RB \\
\emph{do} / VBP $\xrightarrow{\textrm{VC}}$  \emph{have} / VB \\
\emph{do} / VBP $\xrightarrow{\textrm{OBJ}}$  \emph{kids} / NNS \\
\emph{kids} / NNS $\xrightarrow{\textrm{NDET}}$ \emph{any} / DET \\
\end{minipage}

\end{tabular}}
\vspace{\baselineskip}
\caption{{\em (a)} Example of a syntactic derivation tree as can be contained in a treebank (example taken from the Penn Treebank \cite{Marcus:etal:1995}). {\em (b)} Context-free derivation rules induced from the tree in \emph{(a)}. {\em (c)} Example of a dependency tree encoding the syntactic structure of the example in {\em (a-b)}. {\em (d)} Dependency relations induced from the dependency graph in \emph{(c)}. The capitalized symbols are non-terminals and the italicized ones are terminals.}
\label{fig:trees}
\end{figure*}

The task of PCFG induction consists in, given treebank --a collection of parse trees-- inferring the PCFG from which the trees in the treebank were sampled. A simple method for performing this inference is the Maximum-Likelihood method: From each of the internal nodes in a sample of trees, one can infer a context-free derivation rule. The internal nodes in the trees are considered non-terminal symbols, and the leaf nodes in the tree are taken as terminal symbols. Then, if an internal node is labeled $A$, and has as its immediate descendants an ordered list of nodes (leaves or otherwise) $x_1, \ldots, x_k$, with $k \geq 1$, one can infer that the grammar contains the context-free rule $A \to x_1 \ldots x_k$, as is illustrated in the example of Fig.~\ref{fig:trees}. If all trees have the same symbol $A$ as their root, then $A$ is taken to be the starting symbol of the PCFG. If, on the other hand, the trees have different root nodes, one can add a new additional non-terminal symbol $A_0$ as the starting symbol of the PCFG, and also include a context-free rule $A_0 \to A_i$ for each $A_i$ that appears as the root of a tree in the treebank. Finally, the rule probabilities are induced by simple Maximum-Likelihood (ML). If $f[A \to \alpha;T]$ denotes the frequency with which the rule $A \to \alpha$ has been observed in treebank $T$, and $f[A;T]$ is the frequency with which the non-terminal $A$ was encountered, the ML estimate of probability associated to that rule is given by its relative frequency,
\begin{equation}
p \approx \frac{f[A \to \alpha;T]}{f[A;T]} 
\end{equation}
It is known that the PCFG induced by this method is consistent; for an increasingly large treebank, with probability 1, it converges to the true PCFG \cite{Sanchez:Benedi:1997,Chi:Geman:1998} generating it (provided that such one does exist).

\subsection{Derivational Entropy of a PCFG\label{sec:HPCFG}}

The entropy \cite{Shannon:1948} of the parse trees generated by a grammar --PCFG or otherwise-- is given by the usual equation,
\begin{equation}
H[G] = -\sum_{t \in \mathcal{T}[G]} \mathrm{p}_G[t] \log \mathrm{p}_G[t] \label{eq:entropy-trees}
\end{equation}
and is referred to as the {\bf derivational entropy} of the PCFG \cite{Grenander:1967,Grenander:1976}. Notice that, despite being a sum over a potentially infinite set of trees, the magnitude of the entropy remains necessarily finite for the types of grammars that we can be inferred from a treebank by ML  \cite{Sanchez:Benedi:1997,Chi:Geman:1998}. It is therefore a very useful tool for comparing the degree of productivity of two grammars, even when they both potentially produce an infinite number of trees.

PCFGs can potentially generate an infinite number of trees. Therefore, using Eq.~\ref{eq:entropy-trees} directly is most often impractical as it would involve a sum with an infinite number of terms. However, the exact value of $H[G]$ can be computed exactly from a PCFG's production rules \cite{Grenander:1967,Grenander:1976,Miller:OSullivan:1992,Chi:1999} without having to resort to the sum in Eq.~\ref{eq:entropy-trees}.  For this it is necessary to introduce the concepts of the characteristic matrix of a PCFG, and its expansion entropy vectors. 

The {\bf characteristic matrix} of a PCFG is an $\|N\| \times \|N\|$ square matrix $\mathbf{M}_G$ whose $i$-th row -- $j$-th column element $m_{i,j}$ denotes the average number of non-terminals $A_j \in N$ that will be produced in a single expansion of the non-terminal $A_i \in N$ using the rules of the grammar. More formally,
\begin{equation}
\mathbf{M}_G = (m_{i,j}),\quad m_{i,j} = \sum_{r_{i,k} \in R_i} p_{i,k} f(A_j;r_{i,k}) \label{eq:M}
\end{equation}
where $f(A_j;r_{i,k})$ denotes the number of times that $A_j$ occurs in the right-hand side of the rule $r_{i,k}$. In addition, for each non-terminal $A_i \in N$ we define the entropy of the alternative production rules that can expand it, 
\begin{equation}
h_{0}[A_i] = -\sum_{r_{i,j} \in R_i} p_{i,j} \log p_{i,j} \label{eq:local-entropies}
\end{equation}
to which I will refer to as the {\bf local expansion entropy} of symbol $A_i$. If $\mathbf{h}_0$ is the column vector whose components are the $h_{0}[A_i]$ defined above (i.e. the local expansion entropy vector), and $\mathbf{I}$ is the identity matrix of dimension $\|N\|$, the derivational entropies  vector
\begin{equation}
\mathbf{h} = (\mathbf{I}-\mathbf{M}_G)^{-1} \cdot \mathbf{h}_0 \label{eq:inverse}
\end{equation}
has as its components the entropies of the subtrees generated by the grammar, starting by each of its non-terminal symbols. For such entropies to exist, the inverse matrix $(\mathbf{I}-\mathbf{M}_G)^{-1}$ has to exist as well. Fortunately, it is known that for any PCFG  that is proper in the sense of Eq.~\ref{eq:PCFGproper}, the required inverse matrix does indeed exist, and all PCFGs induced from a treebank by ML satisfy this condition \cite{Sanchez:Benedi:1997,Chi:Geman:1998,Chi:1999}. Therefore, if $S=A_k$ is the starting symbol of the grammar, the entropy of Eq.~\ref{eq:entropy-trees} is given by the $k$-th element of its derivational entropies vector $\mathbf{h}$, 
\begin{equation}
H[G] = h_k
\end{equation}

\subsection{Finite Family of a PCFG}
\label{sec:finitefamily}

Any PCFG, can be regarded as the combination of a {\bf skeleton}, $\text{Skel}[G]$, an ordinary (non-probabilistic) context-free grammar $G$ with rules of the form $r \equiv A \to \alpha,\, A \in N, \, (\alpha \in T \cup N)^*$, combined with a {\bf distribution}, $\mathrm{Dist}[G]$, a numerical vector containing the probabilities associated to the individual rules \cite{Soule:1974}. Two PCFGs are said to be {\bf similar} if they share the same skeleton, differing only in their distribution (i.e. they have the same terminal and non-terminal alphabets, and the same rules). For any PCFG we can define the set \cite{Soule:1974},
\begin{equation}
\mathbf{D}_f[G] = \{Dist[G'] \, | \, G'\textrm{ is similar to }G\textrm{ and } \rho(\mathbf{M}_{G'})<1 \}
\end{equation}
where $\rho(\mathbf{M}_{G'})$ denotes the spectral radius (the module of its largest eigenvalue) of the characteristic matrix of $G'$, which by being bounded by 1, guarantees that both MLU and derivational entropies will be finite \cite{Grenander:1967,Grenander:1976,Hutchins:1972,Soule:1974}. I will refer to $\mathbf{D}_f[G]$ as the {\bf finite family} of $G$.

\subsection{Probabilistic Dependency Grammars\label{sec:PDG}}

Context-Free Grammars \cite[CFGs;][of which PCFGs are a probabilistic extension]{Chomsky:1956} build syntax on the concept of {\bf constituency}: The observation that blocks within sentences can be interchanged with other blocks, with the non-terminal internal nodes of the parse trees representing such blocks. An alternative tradition for the description of syntactic structures is that of {\bf Dependency Grammar} \cite[DG;][]{Tesniere:1959}, which relies instead on pairwise dependencies or directed links between words. In this framework, a sentence's syntactic structure is represented by a connected directed graph --which, in most conventions if further restricted to be a directed tree-- with labelled nodes corresponding to either words or broader lexical categories, and vertices denoting their pairwise relations (often also labelled to specify the nature of the relation). These graphs are referred to as dependency graphs. For instance, the syntactic structure of Fig.~\ref{fig:trees}a, could also be represented as the dependency tree in Fig.~\ref{fig:trees}c. In DGs, instead of the context-free productions, the rules of the grammar consist of a list of pairwise directed dependency relations that are licensed by the grammar, such as those in Fig.~\ref{fig:trees}d. Depending on the particular formalism, the relations might be specified between lexical categories and/or, in fully lexicalized models, between individual words. As with the PCFGs, DGs can include probabilities in the dependencies themselves, and are then referred to as Probabilistic Dependency Grammars \cite[PDGs; see, e.g.][for some probability models that can be applied to define a PDG]{Eisner:1996,McDonald:etal:2005}.

\begin{figure*}
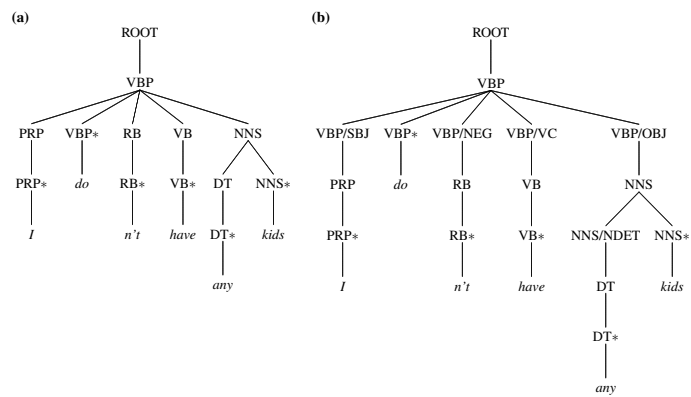

\begin{center}
\resizebox{.6\textwidth}{!}{
\begin{tabular}{ll}
\begin{minipage}[t]{.4\textwidth}
{\bf (a)}\\
\resizebox{\textwidth}{!}{
\synttree[ROOT [VBP [PRP [PRP$*$ [{\em I}]]] [VBP$*$ [{\em do}]] [RB [RB$*$ [\emph{n't}]]] [VB [VB$*$ [{\em have}]]] [NNS [DT [DT$*$ [{\em any}]]] [NNS$*$ [{\em kids}]]]]] 
}
\end{minipage} & 
\begin{minipage}[t]{.6\textwidth}
{\bf (b)}\\
\resizebox{.9\textwidth}{!}{
\synttree[ROOT [VBP [VBP/SBJ [PRP [PRP$*$ [{\em I}]]]] [VBP$*$ [{\em do}]] [VBP/NEG [RB [RB$*$ [\emph{n't}]]]] [VBP/VC [VB [VB$*$ [{\em have}]]]] [VBP/OBJ [NNS [NNS/NDET [DT [DT$*$ [{\em any}]]]] [NNS$*$ [{\em kids}]]]]]]
}
\end{minipage}
\end{tabular}}
\end{center}
\vspace{\baselineskip}
\caption{Representations of the dependency graph in Fig.~\ref{fig:trees}c as a context-free derivation trees: {\bf (a)} Omitting the dependency labels. {\bf (b)} Taking dependency labels into account by means of additional non-terminals.}
\label{fig:treedg}
\end{figure*}

DGs and CFGs are weakly equivalent \cite{Gaifman:1965,Abney:1995} in the sense that they both characterize the set of context-free languages in Chomsky's traditional hierarchy \cite{Chomsky:1956}. Their relation is, however, asymmetric. Any DG can be converted to a strongly equivalent CFG (i.e. with a one-to-one mapping between trees and dependency graphs), but the reverse only holds for a restricted subset of CFGs. For instance, the dependency trees in Fig.~\ref{fig:trees}c can be represented, with or without the dependency labels, by the equivalent parse trees in Fig.~\ref{fig:treedg}. I will make use of this property in order to estimate the syntactic diversity of PDGs.

\subsection{Methods for Correcting Entropy Estimation Bias\label{sec:entropies}}

ML likelihood estimators of entropy from a sample are long known to be biased \cite{Miller:1955}: They will underestimate --sometimes quite severely-- the true value of the entropy of the process that generated the sample. There exists a broad literature on how to correct the bias of entropy estimates obtained from samples, even when these are rather small \cite{Miller:1955,Nemenman:Shafee:Bialek:2002,Chao:Shen:2003,Paninski:2003,Nemenman:Bialek:deRS:2004,Vu:Yu:Kass:2007,Hausser:Strimmer:2009,Chao:etal:2013}. Among these methods, one can distinguish between those that correct entropy estimations among a set of alternatives whose cardinality is either known a priory, or safely assumed to be very low \cite{Miller:1955,Nemenman:Shafee:Bialek:2002,Hausser:Strimmer:2009}, and those that enable corrections for a very large, unknown, and potentially infinite number of alternatives \cite{Paninski:2003,Nemenman:Bialek:deRS:2004,Chao:Shen:2003}. Comparisons on the accuracy of the estimators in this latter group \cite{Vu:Yu:Kass:2007}, reveal that the best performing estimator is the {\bf Coverage-Adjusted Estimator} \cite[{\bf CAE};][]{Chao:Shen:2003}. More recently, the same group that developed the CAE, introduced a new estimator --the {\bf CWJ estimator} \cite{Chao:etal:2013}. This estimator is shown to be less biased and faster converging than the CAE estimator. See Appendix~A for details on these estimators.

\section{The Relation between MLU and Derivational Entropy, and its Implications}
\label{sec:theory}

\subsection{The Crucial Role of MLU}
\label{sec:MLU}

The average length in words of utterances in a corpus, the Mean Length of Utterances (MLU), is by far the oldest measure of syntactic complexity used by researchers. Its first usage, in language acquisition research, dates back to almost a full century ago \cite{Nice:1925}. Still today, it is the most commonly used measure in several fields, including language acquisition, second language learning, and aging research. Despite --or, perhaps, precisely because of-- its simplicity, this measure has some very desirable properties. First, it is very simple the compute as an empirical average, even from unlabelled corpora, and its ML estimator is both consistent and unbiased. Furthermore, it exhibits an extremely fast convergence to its true value, so estimates based on very few samples are already quite reliable \cite{Casby:2011}.
 
Second, the measure is relatively theory-free. It provides an indicator of grammatical complexity that is not bound to any specific theory of syntax. There has been some debate in the literature about the correct units that one should use for measuring MLU, whether one should count words, as in the original measure \cite{Nice:1925}, or one should count morphemes instead \cite{Brown:1973}. In reality, however, the unit of measurement for MLU does not really matter much, for instance, \citet{Parker:Brorson:2005} find both measures to be almost perfectly correlated (Pearson's $r=.99$).

MLU is considered a ``proxy'' measure, in the sense that rather than measuring diversity itself, it measures a quantity that correlates well with it. Some argue that length and syntactic diversity are different things \cite[e.g.][]{Crystal:1974,Klee:Fitzgerald:1985}, and one can find find individual constructions where the two are decoupled. However possible, in actual datasets such occurrences are relatively rare in comparison with the number of cases where MLU and diversity are actually related. At the macroscopic scale, MLU actually correlates very well with many measures of syntactic diversity \cite[e.g.][]{Scarborough:1990,Scarborough:etal:1991,Cheung:Kemper:1992,Norris:Ortega:2009,Agmon:etal:2024}. MLU and derivational entropy are in fact very closely related. As a collateral observation to its main research question, \citet[Fig.~3 on p.~959]{Moscoso:aging:2017} reports that, across the dialogues in the Switchboard I Corpus \cite{SWITCHBOARD}, the relation between derivational entropy and MLU is almost perfectly linear, making both measures virtually indistinguishable (i.e. Pearson's $r=.98$). 

\subsection{Within a PCFG, {MLU} and Derivational Entropy Must Be Directly Proportional}

Such a close relationship between derivational entropy and MLU boils down to a fundamental property of PCFGs.  To see this, let $\boldsymbol{\ell}$ be a vector each of whose $\|N\|$ components is the average length of the strings of terminal symbols that can be directly derived from one of the grammar non terminals. Then, the vector of average length is known \cite{Hutchins:1972,Wetherell:1980} to be,
\begin{equation}
\boldsymbol{\ell} = (\mathbf{I}-\mathbf{M}_G)^{-1} \cdot \boldsymbol{\ell_0} \label{eq:MLU1}
\end{equation}
where $\boldsymbol{\ell_0}$ is a vector whose $i$-th component is the average number of terminal symbols introduced by the rules that directly expand the $i$-th non-terminal symbol (i.e. the non-terminal's MLU):
\begin{equation}
\ell_0[i] = \sum_{r \in R_i} \mathrm{p}[r] \sum_{c \in T} f[c;\mathrm{rhs}[r]] = \mathrm{MLU}[A_i]
\end{equation}
with $R_i$ denoting the set of rules expanding non-terminal symbol $A_i$, $\mathrm{rhs}[r]$  the right-hand side of rule $r$, $\mathrm{p}[r]$ the probability associated with that rule in the PCFG, and $f[c;s]$ the number of times terminal symbol $c$ occurs in string $s \in (T \cup N)^*$. Notice already that Eq.~\ref{eq:MLU1} is identical to Eq.~\ref{eq:inverse} giving a PCFG's derivational entropy, just replacing the entropy vectors $\mathbf{h}$ and $\mathbf{h_0}$ with $\boldsymbol{\ell}$ and $\boldsymbol{\ell_0}$, respectively.

By Eqs.~~\ref{eq:inverse} and ~\ref{eq:MLU1}, the $i$-th row vector of matrix $(\mathbf{I}-\mathbf{M}_G)^{-1}$, which I will abbreviate to $\mathbf{v}_i$, determines both the length of the strings generated by non-terminal $A_i$ and the entropy of the subtrees it generates. With this notation, the entropy and MLU generated by a grammar are given by plain dot products:
\begin{eqnarray}
H[A_i]  = h[i] & = \mathbf{v}_i \cdot \mathbf{h_0} & = | \mathbf{v}_i | \, | \mathbf{h_0} | \, \cos[\mathbf{v}_i,\mathbf{h_0}] , \nonumber \\
\mathrm{MLU}[A_i] = \ell[i] & = \mathbf{v}_i \cdot \boldsymbol{\ell_0} & = | \mathbf{v}_i | \, | \mathbf{\boldsymbol{\ell_0}} | \, \cos[\mathbf{v}_i,\boldsymbol{\ell_0}] , \label{eq:dotprods}
\end{eqnarray}
where $| \mathbf{x} |$ denotes the module of vector $\mathbf{x}$. All vectors $\mathbf{h_0}$, $\boldsymbol{\ell_0}$, all $\mathbf{v}_i$, and --as a consequence-- both the cosines are non-negative. This last point is certain because, for all grammars induced from a treebank by ML, $\mathbf{M}_G$ is non-singular and its largest eigenvalue is positive  and smaller than one, $\rho(\mathbf{M}_G)<1$ \cite{Sanchez:Benedi:1997,Chi:Geman:1998,Chi:1999}. In turn, this entails that $(\mathbf{I}-\mathbf{M}_G)$ is an M-matrix \cite{Ostrowski:1937}, whose inverse is always non-negative. Dividing both components of Eq.~\ref{eq:dotprods}:
\begin{equation}
\frac{H[A_i]}{\text{MLU}[A_i]} = \frac{|\mathbf{h_0}|}{|\boldsymbol{\ell_0}|} \, \frac{\cos[\mathbf{v_i},\mathbf{h_0}]}{\cos[\mathbf{v_i},\boldsymbol{\ell_0}]} = \alpha_i \geq 0 \label{eq:ratio1}
\end{equation}
In sum, the MLUs and derivational entropies of the non-terminals of a grammar are the result of a common, non-negative linear transformation. Therefore, one should expect a \emph{grammar-internal} correlation such that
\begin{equation}
H[A_i]  = \alpha_i \, \mathrm{MLU}[A_i] \label{eq:corr-internal}
\end{equation}
The relation in Eq.~\ref{eq:corr-internal} is not an approximation, but a mathematical identity. Computing these two measures for the root symbol of the grammar indicates that MLU is not just a proxy any longer, but an explicit --theoretically motivated-- measure of syntactic diversity. Furthermore, for realistic grammars with more than about 20 non-terminals, the ratio of cosines in Eq.~\ref{eq:ratio1}, with all three vectors in the positive hyper-quadrant, is expected to approach 1.0 with convergence as a function of $\|N\|$ (the number of non-terminals, which is also the maximum possible dimensionality of the space). In simulations I found the standard deviation from one to be less than $1/\sqrt{\|N\|}$). As a consequence within a PCFG we should expect an extremely strong positive correlation between the MLUs and derivational entropies across the nonterminals. In other words, MLU and derivational entropy are expressing the same, with a mere change in units of measurement.

\subsection{The Derivational Entropy Rate}
\label{sec:deventrate}

The proportionality in Eq.~\ref{eq:corr-internal} suggests that one can define the {\bf derivational entropy rate} of a grammar as the average increase in derivational entropy resulting from adding an individual symbol to a string using the grammar:
\begin{equation}
h[G] = \frac{H[G]}{MLU[G]} = \alpha > 0 \label{eq:rate1}
\end{equation} 

The observation of \citet{Moscoso:aging:2017} goes one step further than mere proportionality. The telephone conversations contained in the Switchboard I Corpus are diverse with respect to the topic of conversation, the socio-cultural origins, ages, and sex of the speakers. Nevertheless, the very strong correlation between the MLU and derivational entropies of the files would suggest that \emph{the derivational entropy rate is roughly constant across all conversations in the corpus}. What all these conversations share is that they all happened in (different sub-dialects of) spoken American English and that they were all parsed using the same grammatical annotation convention, that of the Penn Treebank \cite{Marcus:etal:1995}. This motivates putting forward two hypotheses of decreasing generality:
\begin{description}
\item [Hypothesis 1:] The derivational entropy rate of a PCFG is constant across PCFGs representing the same or closely related languages.
\item [Hypothesis 2:] The derivational entropy rate of a PCFG is constant across PCFGs representing the same or closely related languages, annotated using the same grammatical convention.
\end{description}
In Section~\ref{sec:corpusan}, I will test the validity of these  hypotheses.

\subsection{Relativity of Derivational Entropy}
\label{sec:relativity}

Consider now two PCFGs on the same terminal alphabet, $G_1 = (T, N_1=\{A_1,\ldots,A_{\|N_1\|}\},A_1,R_1)$ and $G_1 = (T, N_2=\{B_1,\ldots,B_{\|N_2\|}\},B_1,R_2)$ that generate the same probabilistic language ($\mathcal{L}(G_1) = \mathcal{L}(G_2)$) or, that at least generate strings of the same average length. These two PCFG can, for instance, be ML-induced grammars from a context-tree treebank and a dependency treebank annotating the same corpus. MLU is the same for both grammars:
\begin{equation}
\mathrm{MLU}[A_1] = \mathrm{MLU}[B_1] \label{eq:external1}
\end{equation}
Both PCFGs are subject to the grammar internal correlation of Eq.~\ref{eq:corr-internal},
\begin{align}
H[A_i] & = \alpha_1 \mathrm{MLU}[A_i]\nonumber \\
H[B_i] & = \alpha_2 \mathrm{MLU}[B_i] \label{eq:external2}
\end{align}
with $\alpha_1, \, \alpha_2 > 0$. Combining Eq.~\ref{eq:external1} and Eq.~\ref{eq:external2}, one can express the \emph{external} relationship between the derivational entropies of both grammars in terms of the parameters of their internal entropy-length relations,
\begin{equation}
H[G_2] = \frac{\alpha_2}{\alpha_1} H[G_1] = \beta \, H[G_1], \qquad \beta \geq 0\label{eq:external3}
\end{equation}
Hence, we can expect that the entropies of grammars describing the same languages will be directly proportional to each other (without an intercept).

Suppose we have multiple corpora of the same language annotated according to two different grammatical theories. The actual entropy values of the corpora will change depending on which of the two grammatical theories was used for parsing. However, as long as the grammatical theories are sensible, it should be expected that --in the vast majority of cases-- different grammatical structures will in most cases receive different parses in either theory, and identical grammatical structures will in most cases remain identical in both theories. This implies that the relative values of the entropies for different corpora should be strongly correlated across the two theories. For this to be true, the value of parameter $\beta$ needs to be roughly constant across pairs of treebanks parsed according to two different conventions. Interestingly, this is exactly what one would predict if Hypothesis~2 above were true: The derivational entropy rates ($\alpha_1$ and $\alpha_2$) would be constant for a particular grammatical annotation convention, and so would $\beta$. In other words, one would expect that grammatical entropy estimates from different corpora are, \emph{in relative terms}, independent of the grammatical convention used in building the treebanks: If corpus A has a greater diversity than corpus B according to grammatical theory 1, one should expect that, Hypothesis~2 implies that the diversity of corpus A is also greater than that of B. On the other hand, Hypothesis~1 would have even stronger implications, the derivational entropy rates in both corpora should be identical (because they are dealing with the same language) and hence the derivational entropies should also be identical. These implications will be tested in Section~\ref{sec:corpusan}.

\section{Measuring Derivational Entropy from Small Treebanks}
\label{sec:SITEsec}

\subsection{Generalizing from Induced PCFGs}

If one's goal is to estimate the derivational entropy of the grammar from which a particular corpus was sample, rather of considering just the directly observed syntactic trees, one should first generalize over the structures observed in the corpus. The goal is to account not just for the directly observed structures, but also for those that could have occurred. A way to achieve this is by inducing a grammar, and then estimating the entropy from the induced grammar, rather than from the original treebank. Approaches similar to this have been followed by several studies \cite{Hale:2003,Hale:2006,Corazza:etal:2013,Linzen:Jaeger:2014,Moscoso:2014,Moscoso:aging:2017}. The additional grammar induction step ensures that the internal structure of the parses is explicitly considered: Evidently, the induced grammar accounts for all the syntactic parses that have been directly observed. In addition to these, it also takes into account infinitely many other parses that, despite not having been actually found on the treebank, could be built by recombination of the partial structures observed in the corpus:
\begin{equation}
\underbrace{t_1, t_2, \ldots, t_{\|T\|}}_{\text{Observed structures}}, \underbrace{t_{\|T\|+1}, t_{\|T\|+2}, t_{\|T\|+3}, \ldots}_{\text{Generalized  plausible structures}}
\end{equation}
By generalizing, the grammar is very likely to also generate many structures that, on human inspection, can be determined to be actually impossible. This last point is less of an actual problem than one might think. What we are interested in measuring is the degree to which the observed parses provide evidence for internal structure. The over-productivity of the induced grammars, from this perspective, is just a tool for measuring internal structure, rather than an explicit model for the precise structures that an individual might actually produce. Evidently, the latter goal would require to model well beyond syntax, including explicit models of semantics, pragmatics, and all kinds of socio-cultural aspects. In short, we are concerned with measuring syntactic diversity on a macroscopic scale, rather than identifying individual syntactic structures at a microscopic one.

\subsection{ML Estimator\label{sec:MLE}}

The simplest way for inducing a grammar from a treebank is using the ML method described in Section~\ref{sec:MLPCFG}. Once such a grammar has been induced, there are some options on how to estimate its entropy. The most evident one would be to just compute the maximum-likelihood approximations of the characteristic matrix ($\mathbf{M}_G^\mathrm{ML}$) and local expansion entropy vector ($\mathbf{h_0^{\mathrm{ML}}}$) directly from the induced grammar, so that the ML estimator for the entropy can be computed by simply applying Eq.~\ref{eq:inverse} to compute the ML estimate of the derivational entropies vector:
\begin{equation}
\mathbf{h}^\mathrm{ML} = (\mathbf{I} - \mathbf{M}_G^\mathrm{ML})^{-1} \cdot \mathbf{h_0^{\mathrm{ML}}}\label{eq:MLE}
\end{equation}
whose $k$-th component (if $A_k$ is the root symbol of the PCFG) is the entropy estimator:
\begin{equation}
H^\mathrm{ML}[G] = h^\mathrm{ML}_k\label{eq:ML2}
\end{equation}
This was the method used for computing derivational entropies by several previous studies \cite{Hale:2003,Hale:2006,Linzen:Jaeger:2014}.

\subsection{ML-Monte Carlo Estimator\label{sec:MCE}}

An interesting alternative to this ML method was used by \citet{Corazza:etal:2013}. Using a sampling approach to ML estimation, they take the true probability of the trees ($\mathrm{p}_G[t]$) to be unknown, but approximated by the probability of the tree using the  grammar induced by ML from a training subset of the treebank ($\mathrm{p}_G^\mathrm{ML}[t]$). An upper bound for the true entropy of the underlying grammar would be provided by the cross-entropy \cite{Shannon:1948} between both distributions:
\begin{equation}
H[G\|\mathrm{ML}] = -\sum_{t \in \mathcal{T}(G)}  \mathrm{p}_G[t] \log \mathrm{p}_G^\mathrm{ML}[t] \geq H[G] \label{eq:falseML}
\end{equation}
As the true tree probabilities ($\mathrm{p}_G$) are unknown, \citet{Corazza:etal:2013} take advantage that, by the Shannon-McMillan-Breiman theorem \cite{Breiman:1957,Chung:1961}, the treebank itself can be considered a sample from the true $\mathrm{p}_G$ that will converge with a sufficiently large corpus. Therefore, they propose using a Monte Carlo estimator for Eq.~\ref{eq:falseML}, by sampling trees from a testing subset of the treebank ($\mathrm{p}_T[t]$) as an approximation of the true $\mathrm{p}_G[t]$,  and hence one can use this testing subsample for computing the estimated cross-entropy:
\begin{equation}
\hat{H}[G\|\mathrm{ML}] = -\sum_{t \in T} \mathrm{p}_T[t] \log \mathrm{p}_G^\mathrm{ML}[t] = -\sum_{t \in T} \frac{f[t;T]}{\|T\|} \log \mathrm{p}_G^\mathrm{ML}[t] \label{eq:Montecarlo}
\end{equation}
This is indeed a consistent estimator for the cross-entropy in the sense that, for an infinite corpus size it would be converge on the true cross-entropy of the system. Furthermore, in such limit condition, the cross-entropy itself would actually converge on the true value of the the entropy (i.e. it would not even be an upper bound, but an exact estimator), because both probability estimators $\mathrm{p}_T$ and $\mathrm{p}_G^\mathrm{ML}$ would converge on $\mathrm{p}_G$ given an infinitely large treebank. It is, however, incorrect that the estimated $\hat{H}[G,\mathrm{ML}]$ would be an upper bound of the true entropy value. Rather than a cross-entropy --which would indeed be an upper bound-- this is a sample-based ML estimator of the cross-entropy, an it is subject to exactly the same estimation bias \cite{Miller:1955} as the entropy: It will be an underestimation. Furthermore, note that both  $\mathrm{p}_T$ and $\mathrm{p}_G^\mathrm{ML}$ are ML approximations of $\mathrm{p}_G$ taken from two subsamples from the same population. One therefore should expect that $\mathrm{p}_T \propto \mathrm{p}_G^\mathrm{ML}$, and as a direct consequence, the cross-entropy estimator is in fact Monte Carlo estimator of the ML entropy:
\begin{equation}
\hat{H}[G\|\mathrm{ML}] = H^\mathrm{MC}[G] \approx H^\mathrm{ML}[G]
\end{equation}
Indeed, as we will see in the corpus analyses below, the estimators in Eq.~\ref{eq:ML2} and Eq.~\ref{eq:Montecarlo} are virtually indistinguishable, and are both underestimations of the true entropy. 

\subsection{Smoothed Induced Treebank Entropy (SITE)\label{sec:SITE}}

The crucial insight of the SITE measure is that the estimate of the PCFG entropy can be decomposed into $\|N\|$ local expansion entropy estimates. When inducing a grammar by ML from a treebank $T$, both the characteristic matrix ($\mathbf{M}_G^\mathrm{ML}$) and the local expansion entropies ($\mathbf{h_0}^\mathrm{ML}$) are ML estimates of their true values. On the one hand, the elements $m_{i,j}$ of $\mathbf{M}_G^\mathrm{ML}$ are counts of the average number of instances of non-terminal $A_i$ that result from expanding an instance of non-terminal $A_i$. These counts should converge very fast; they are plain frequency estimates that will be unbiased even when estimated from quite small treebanks, and therefore do not need correction. Furthermore, by not modifying the characteristic matrix, one ensures that its spectral radius remains bounded by one as the result of ML PCFG induction always is \cite{Sanchez:Benedi:1997,Chi:Geman:1998,Chi:1999}, and hence the derivational entropies will also converge after correction \cite{Grenander:1967,Grenander:1976}. On the other hand, the expression for the local expansion entropies (Eq.~\ref{eq:local-entropies}) is just the classical entropy equation. The $h_{T,0}[A_i]$ values are ML estimates of the uncertainty on the different ways in which symbol $A_i$ can be expanded. As with any other ML entropy estimate, these should be expected to be negatively biased. This bias, however, is of the precise type that can be corrected with the sort of estimators given in Section~\ref{sec:entropies}.

\citet{Moscoso:2014} used such an approach. In order to estimate the entropy of the grammar from which a treebank was sampled, PCFGs were induced from the treebanks, and then used the plain ML estimates of the characteristic matrices, together with the CAE of the local expansion entropy vectors for computing the derivational entropy of the grammars (comparing samples originating in different historical periods). \citet{Moscoso:aging:2017} also followed this approach, but using the CWJ estimator instead of the CAE for smoothing the local expansion entropies. As I will demonstrate with corpus analyses, SITE with CWJ local entropy smoothing achieves extremely fast, unbiased convergence to the true values of the derivational entropy associated with a treebank.

\subsection{SITE on Dependency Treebanks}

SITE can also be computed to dependency treebanks. For doing this, one just needs an additional preprocessing step in which all dependency graphs in the treebank are converted to equivalent context-free derivation trees, as is illustrated in the change from the dependency structure in Fig.~\ref{fig:trees}c into the derivation tree in Fig.~\ref{fig:treedg}b.

\begin{figure}[!h]
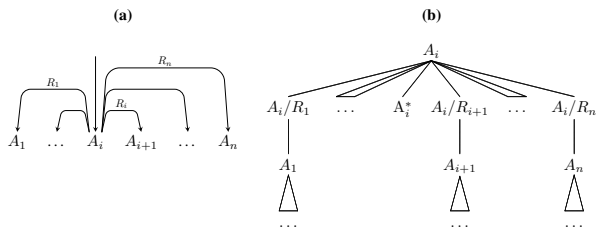

\begin{center}
\resizebox{.5\textwidth}{!}{
\begin{tabular}{cc}
{\bf (a)} & {\bf (b)} \\
\savecellbox{\begin{dependency}[text only label, label style={above}]
\begin{deptext}[column sep=0.3cm]
$A_1$ \& \dots \& $A_i$ \& $A_{i+1}$ \& \ldots \& $A_n$ \\
\end{deptext}
\depedge{3}{1}{$R_1$}
\depedge{3}{2}{}
\depedge{3}{4}{$R_i$}
\depedge{3}{5}{}
\depedge{3}{6}{$R_{n}$}
\deproot{3}{}
\end{dependency}}&
\savecellbox{\synttree{2}[$A_i$ [$A_i/R_1$ [$A_1$ [.x $\ldots$]]] [.x $\ldots$] [$\mathrm{A}^*_i$] [$A_i/R_{i+1}$ [$A_{i+1}$ [.x $\ldots$]]] [.x $\ldots$] [$A_i/R_{n}$ [$A_n$ [.x $\ldots$]]]]} \\
\\[-\rowheight] \printcelltop & \printcellmiddle
\end{tabular}}
\end{center}
\caption{Expansion of node $A_i$ with the relations depicted in \emph{(a)} into the corresponding part context-free derivation tree in \emph{(b)}.}
\label{fig:expansion}
\end{figure}

The procedure for converting a dependency graph into a derivation tree is a variation of the procedure described by \citet{Gaifman:1965}, with the addition of nodes for representing the dependency labels. Variants of this procedure have also also been used in recent studies \cite[e.g.][]{Rezaii:etal:2022}
For a dependency graph with $n$ nodes $A_1, \ldots, A_n$, and $n-1$ labelled relations $R_1$, $R_2$, \ldots, $R_{n-1}$, one creates a context-free derivation tree with $2n$ internal nodes. Of these, $n$ correspond to the nodes in the original dependency graph (labelled $A_1, \ldots, A_n$), one is a root node $A_0$, and  $n-1$ correspond to the relations (labelled $A_i / R_j$, denoting a relation with label $R_j$ going from node $A_i$ to node $A_j$ with label $R_j$). In addition, the tree has $n$ leaves (labelled $\mathrm{A}^*_1, \ldots, \mathrm{A}^*_n$). The root node of the dependency graph is made to hang from the tree root $A_0$. Then for each node in the dependency graph, a corresponding context free branching is constructed, by making all the relations originating from it into daughters in the context-free graph, from which the corresponding nodes in the relation are made to hang. In addition, each internal node $A_i$ has a leaf daughter $A^*_i$, positioned in its corresponding order in the original dependency graph.The process --illustrated in Fig.~\ref{fig:expansion}-- is then recursively repeated for each of the symbols that depend from $A_i$. For projective dependency trees, this procedure is a reversible one in the physical sense: The derivation trees constructed by this method preserve all information that was encoded in the original dependency structure. After applying this preprocessing to the corpus, a PCFG can be induced by the usual method, and its entropy can then be estimated by SITE.

\section{Corpus Analyses}
\label{sec:corpusan}

\subsection{Accuracy of the SITE Method for Corpora with Known Entropy}
\label{sec:c1}

In order to assess the accuracy and bias reduction of the SITE method in estimating the entropy of  realistic treebanks with known true entropy value, I used the Wall Street Journal subsample of the Penn Tree Bank \cite{Marcus:etal:1995} distributed with Python's Natural Language Toolkit \cite[NLTK v3.9.1;][]{nltk}. This sample includes 3,914 parsed sentences. Importantly, the same sentences are available both as context-free derivation trees, and as dependency graphs. I induced a PCFG grammar from each version of the treebank. These grammars were used as the true underlying grammars, from which the task was estimating the true value of the derivational entropy.

This known entropy case is useful when evaluating the properties or ambiguity of existing grammars, parsing, and generation systems. In Section~\ref{sec:c2} I will study the more common case, where the actual true grammars are unknown, and so are their true derivational entropy values.

\subsubsection{Context-Free Treebank Case}
\label{sec:c1a}

The context-free trees were pre-terminalized by removing the leaves of the original trees (corresponding to the actual words), so that the leaves of the transformed tree were part-of-speech (POS) tags. The PCFG induced by ML from this corpus contains a total of 8,009 context-free production rules using an alphabet of 662 non-terminal symbols, and it has a true entropy value of 103.445 bits (i.e. the equivalent of generating $1.38\cdot10^{31}$ distinct equiprobable derivation trees).

From the induced PCFG, I sampled artificial corpora of 24 increasing sizes, from 1 to 15,000 sentences (this number was chosen as it was found that this is how much the ML method required for converging on the true entropy value for this dataset), with the subsample sizes evenly spaced on a logarithmic scale (i.e. sample sizes of 1, 2, 3, 5, 7, 11, 17,  25,  37, 55, 82, 122, 183, 273, 407, 608, 908, 1,355, 2,023, 3,020, 4,509, 6,731, 10,048, and 15,000 sentences). From each of the generated corpora, I induced a PCFG and estimated their entropies using the plain ML \cite[as in][]{Hale:2003,Hale:2006,Linzen:Jaeger:2014}, and Monte Carlo ML \cite[as in][]{Corazza:etal:2013}, and the SITE methods, using either the CAE \cite[as in][]{Moscoso:2014} or CWJ \cite[as in][]{Moscoso:aging:2017} entropy smoothers from each of the samples. This process was repeated one hundred times.

\begin{figure*}[th]
\begin{center}
\begin{tabular}{cc}
{\bf (a)} & {\bf (b)} \\
\includegraphics[width=.3\textwidth]{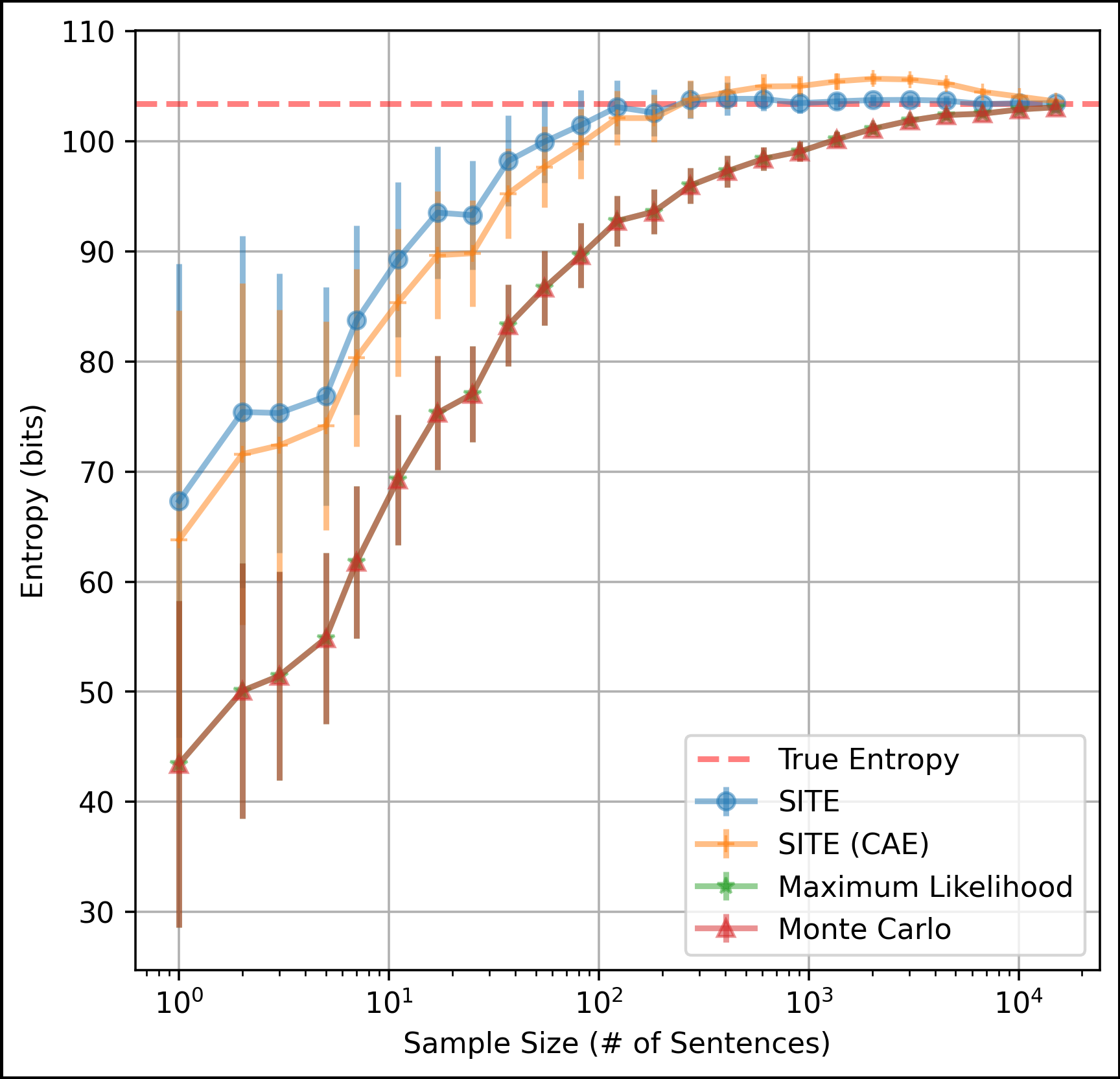} &
\includegraphics[width=.3\textwidth]{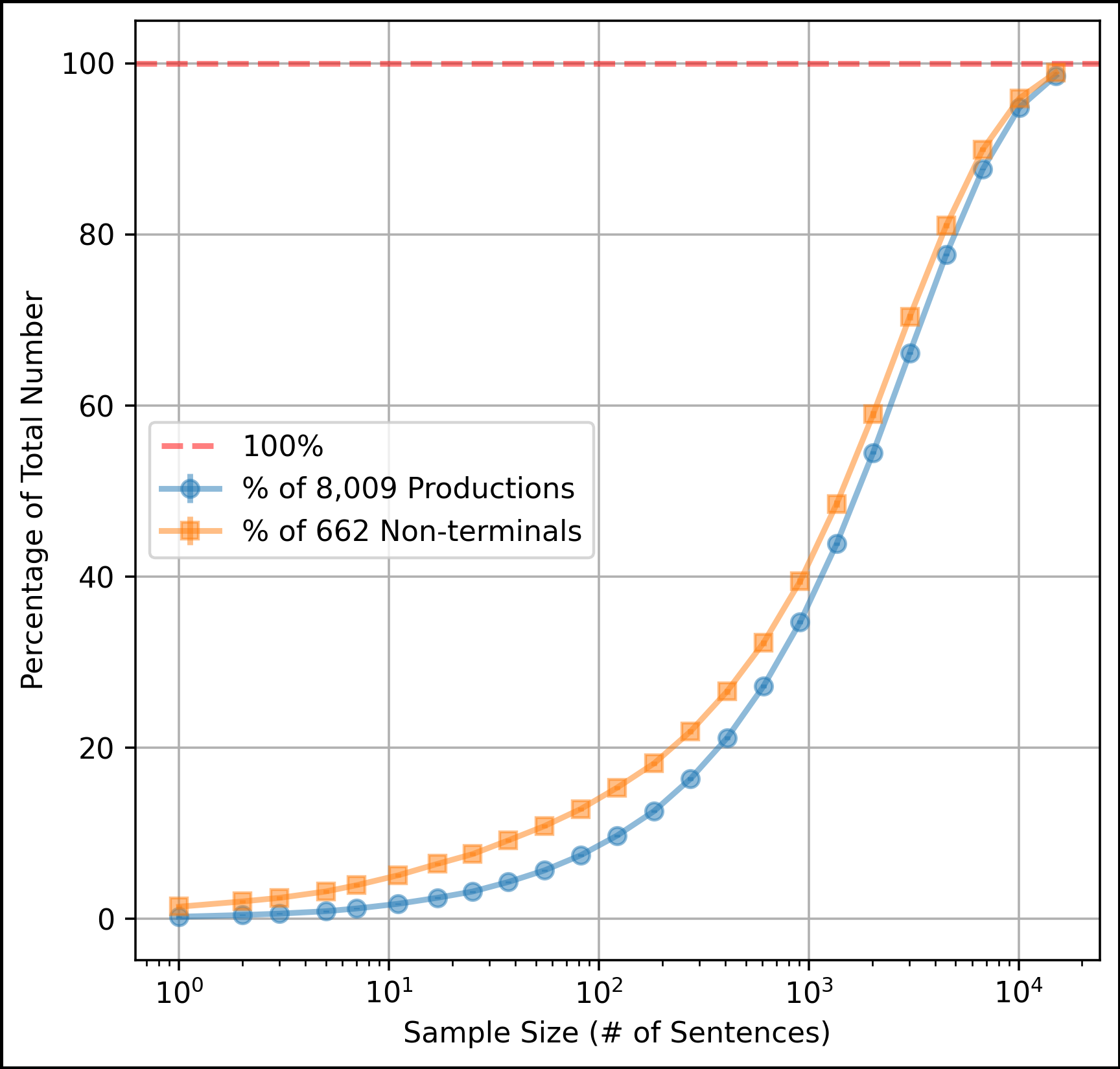}
\end{tabular}
\end{center}
\caption{Results of Corpus Analysis~\ref{sec:c1a}. Error bars are 95\% confidence intervals for the mean. Notice the horizontal logarithmic scales. {\bf (a)} Estimated derivational entropies according to the four estimators as a function of the sample size. {\bf (b)} Percentage of rules and non-terminal symbols from the original grammar that are documented as a function of sample size (the 95\% C.I.s are imperceptible).}
\label{fig:sim1}
\end{figure*}

Fig.~\ref{fig:sim1} plots the resulting entropy estimates. Panel {\bf (a)} compares the accuracy of the three estimation methods for increasing sample sizes. Notice that both versions of the SITE method converge must faster than the ML estimates (which are indistinguishable from each other). Both SITE estimates approach the true value with just 122 sentences, and with just 55 sentences the true entropy value was within the 95\% confidence interval for the mean estimate, something that took about 15,000 sentences to achieve with either of the ML methods (which are indishtinguishable). In more extreme terms, the ML estimates with over one hundred sentences are about as biased as is the SITE estimate with a single sentence. In terms of overall error, the CS and CWJ methods achieve very similar performances. However, the CS-based method slightly overshoots the entropy estimates, and in practice is as biased (but in opposite direction) as  the ML methods are for samples of over 3,020 sentences. In contrast, the SITE method using the CWJ estimates completely eliminates the bias from 120 sentences onwards. Panel {\bf (b)} illustrates how the ML methods do not achieve true convergence until  every non-terminal symbol and production rule from the true grammar have been explicitly observed in the corpus.

\subsubsection{Dependency Treebank Case}
\label{sec:c1b}

Each of the dependency graphs in the corpus was converted to a context-free derivation tree (the dependencies were considered as relating POS tags rather than words). From the resulting derivation trees, I induced a PCFG by ML, obtaining a PCFG with 8,104 context-free production rules using an alphabet of 46 non-terminal symbols, and a true entropy value of 78.36 bits (i.e. the equivalent of generating $3.88 \cdot 10^{23}$ distinct equiprobable dependency trees; notice that the dependency paradigm is, for equivalent corpora, substantially more restrictive in its generalization than the context-free paradigm). As was done in Corpus Analysis~\ref{sec:c1a}, from the estimated PCFG, I sampled artificial corpora of the same 24 increasing sizes. The entropies of the induced PCFG were estimated using both ML methods, and two variantes of SITE. This process was repeated one hundred times.

\begin{figure*}[th]
\begin{center}
\begin{tabular}{cc}
{\bf (a)} & {\bf (b)} \\
\includegraphics[width=.3\textwidth]{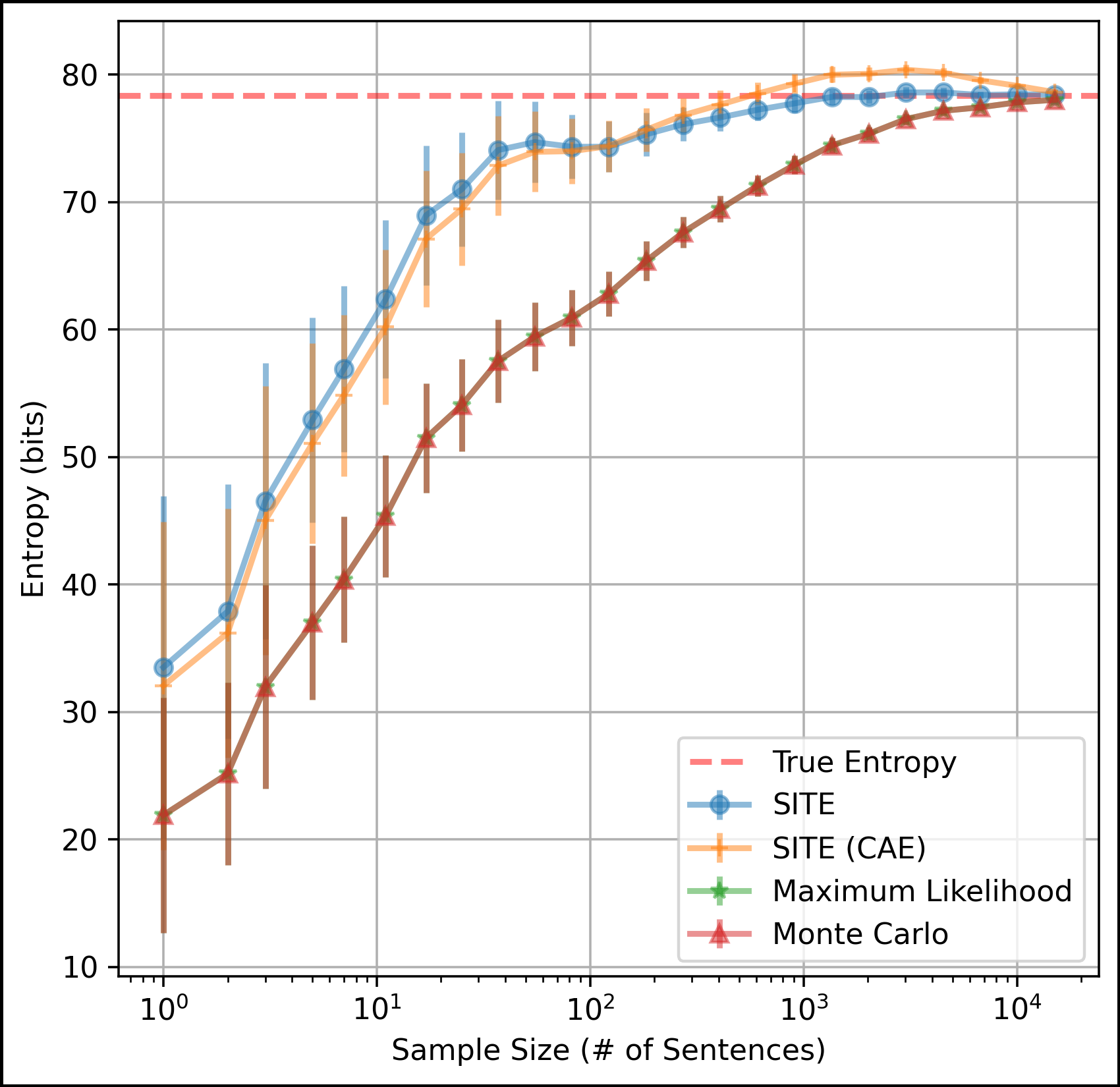} &
\includegraphics[width=.3\textwidth]{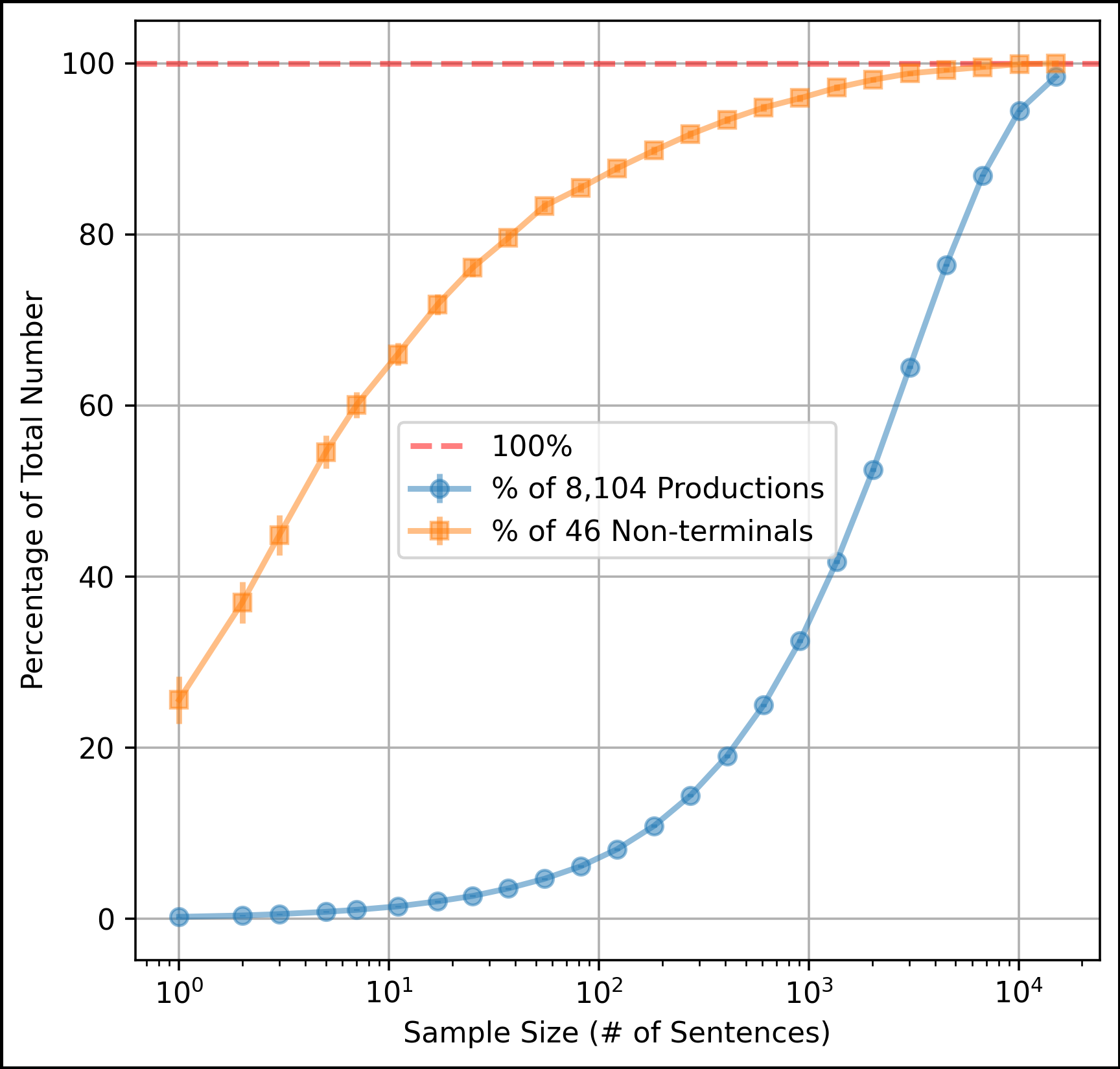}
\end{tabular}
\end{center}
\caption{Results of Corpus Analysis~\ref{sec:c1b}. Error bars are 95\% confidence intervals for the mean. Notice the horizontal logarithmic scales. {\bf (a)} Estimated derivational entropies according to the four estimators as a function of the sample size. {\bf (b)} Percentage of rules and non-terminal symbols from the original grammar that are documented as a function of sample size (the 95\% C.I.s are imperceptible).}
\label{fig:sim2}
\end{figure*}

Fig.~\ref{fig:sim2} plots the resulting entropy estimates. Panel {\bf (a)} compares the accuracy of the three estimation methods for increasing sample sizes. Notice that both versions of the SITE method converge must faster than the plain ML estimate, both approaching the true value with just 608 sentences, something that --as before-- took about 15,000 sentences to achieve with the plain ML methods. In terms of overall error, the CS and CWJ methods once more achieve very similar performances, with the CS method again slightly overestimating the entropies. The SITE method using the CWJ estimates are very accurate from about 407 sentences, and it completely removes the bias from 908 sentences onwards. As before, Panel {\bf (b)} shows that the ML methods do not achieve convergence until all rules and non-terminals of the grammar have been directly observed.

\subsubsection{MLU-Derivational Entropy Correlations}
\label{sec:c1c}

In order to investigate whether the derivational entropy rate is relatively constant for a given grammatical theory, I compare the entropy estimates for the context-free and dependency versions of the samples used in Corpus Analyses~\ref{sec:c1a} and~\ref{sec:c1b} . These corpora are subdivided into 199 very small files, ranging in sizes from just a single sentence, to 185 of them (mean size $19.67 \pm 1.64$ sentences/file). For each of these files, I estimate the MLU, and derivational entropy using SITE (based on pre-terminalized parse trees and dependency graphs), according to both versions of the corpora. 

\begin{figure*}[th]
\begin{center}
\begin{tabular}{cc}
{\bf (a)} & {\bf (b)} \\
\includegraphics[width=.3\textwidth]{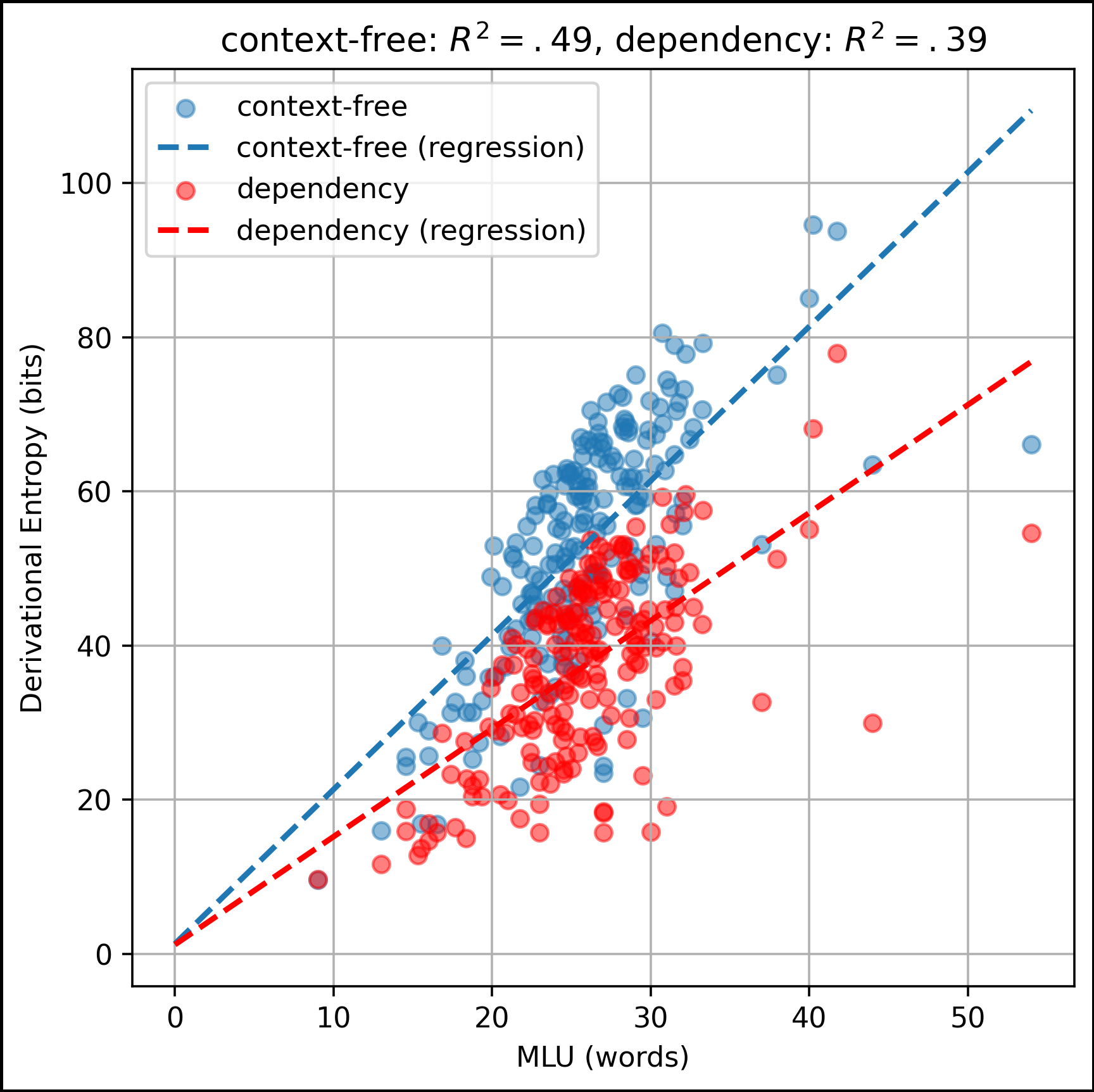} &
\includegraphics[width=.3\textwidth]{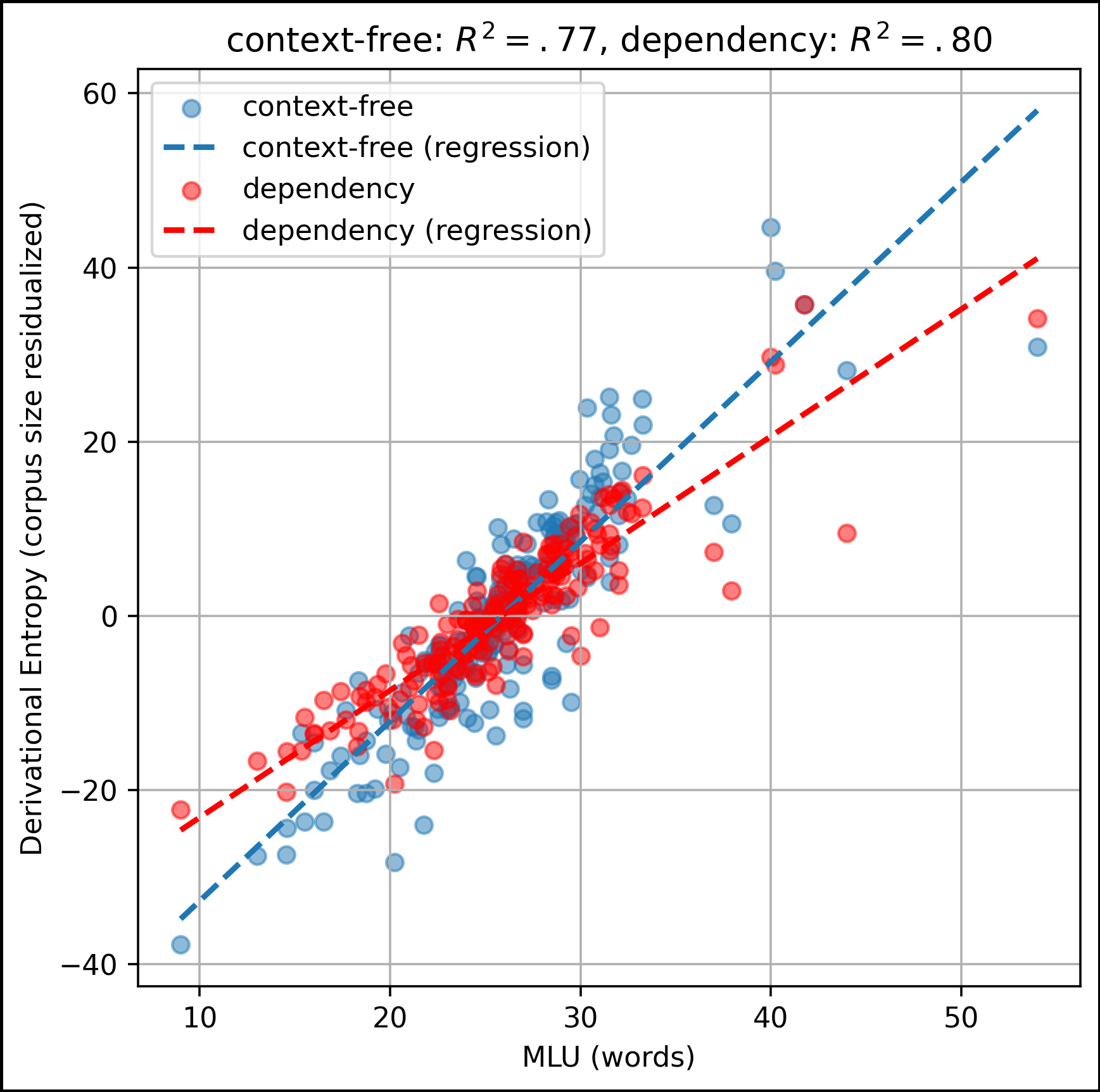}
\end{tabular}
\end{center}
\caption{Results of Corpus Analysis~\ref{sec:c1c}. Each point plots the derivational estimates of a file as a function of the MLU, in the context-free (blue) and dependency (red) versions of the treebank. The dashed lines plot linear regressions. {\bf (a)} Relationship between MLU and the raw derivational entropies. {\bf (b)} Relationship between the MLU and the derivational entropies after residualizing log corpus size.}
\label{fig:sim3MLU}
\end{figure*}

Panel {\bf (a)} in Fig.~\ref{fig:sim3MLU} plots the derivational entropies for each subcorpus as a function of their MLUs, for both the context-free (blue dots) and dependency (red dots) versions. There are clear correlations between the measures (Pearson's $r=.70,p<.0001$ for context-free versions, and Pearson's $r=.63,p<.0001$ for the dependency versions). Furthermore, both regression lines (dashed lines in the plot) converge almost exactly at the origin of coordinates, indicating the absence of any significant intercept (i.e. the regressions were fitted allowing for an intercept to be present, but they were insignificant; $1.29\pm3.81, t[197]=.34, p=.73$ for context-free and $1.18\pm3.29, t[197]=.359, p=.72$ for dependencies). This is in full agreement with the prediction of Eq.~\ref{eq:rate1}, of MLU and derivational entropy being directly proportional. From the same equation, it follows that the regression slopes estimate the derivational entropy rates. If we refit the regressions to exclude the non-significant intercepts, the estimated derivational entropy rates are $h[G] \approx 2.05 \pm .03$ bits/symbol ($t[198]=70.88, p<.0001$) for the context-free paradigm treebank, and $h[G] \approx 1.44 \pm .02$ bits/symbol ($t[198]=57.84, p<.0001$) for the dependency treebank.

Although the regressions are exactly as we predicted, the overall correlation coefficients are nowhere as strong as the Pearson's $r=.98$ reported by \citet{Moscoso:aging:2017}. Notice, however, that for such small corpora (the majority of which were well below one hundred sentences) even the fast-converging SITE is most likely severely underestimating the derivational entropies (the MLUs are, on the other hand, unbiased estimators). As illustrated by Figs.~\ref{fig:sim1} and \ref{fig:sim2}, for very small corpora there is an approximately logarithmic relation between corpus size and the derivational entropy estimates, even after applying the SITE corrections. We can therefore correct the derivational entropy estimates by residualization: I take the residuals from a linear regression predicting the value of the derivational entropy as a function of the log corpus size. After this residualization, there wasn't any additional non-linear relation between corpus size and either of the two residualized entropies, as revealed by the non-significant rank correlations (Spearman's $\rho=.05, p=.52$ for the context-free entropy, and Spearman's $\rho=.05, p=.44$ for the dependency entropy). Panel {\bf (b)} in Fig.~\ref{fig:sim3MLU} shows how, when the estimation biases are removed by residualization, the correlations between MLU and derivational entropy indeed become much stronger (Pearson's $r=.88, p<.0001$ for the context-free corpora, and Pearson's $r=.89, p<.0001$ for the dependency corpora), much more in line with the almost perfect relationship reported by \citet{Moscoso:aging:2017}. It indeed appears that the derivational entropy rates are roughly constant across the 199 subcorpora, depending just on which paradigm was used for annotating the parses. The difference in the value of the derivational entropy rate between the DG and CFG versions of the same corpus falsifies Hypothesis~1 of Section~\ref{sec:deventrate}, that the constancy of the derivational entropy rate is a consequence of texts being in the same language. Even for identical texts, these entropy rates may change as a result of using different conventions for corpus annotation, as are in this case the CFG and DG syntactic annotation paradigms. These results, however, remain consistent with Hypothesis~2, that the derivational entropy rate remain constant within a language and annotation convention.

\subsubsection{Relative Independence of Grammatical Formalism}
\label{sec:c1d}

The previous analyses found the derivational entropy rates to be roughtly constant within a grammatical paradigm, consistent with Hypothesis~2 of Section~\ref{sec:deventrate}. As argued in Section~\ref{sec:relativity}, this in turn predicts that the derivational entropies are indeed relatively independent of the grammatical formalism used for the treebank. By Eq.~\ref{eq:external3}, the dependency based derivational entropies should be directly proportional to the context-free ones by a factor equal to the ratio of their derivational entropy rates. Taking the values estimated in the previous analysis, the ratio between the two derivational entropy rates is expected to be:
\begin{equation}
\beta \approx \frac{1.44}{2.05} = .70
\end{equation}
In other words, the derivational entropy of a the dependency annotated corpus should be about 70\% of the derivational entropy of the same corpus annotated using the context-free paradigm.

\begin{figure*}[th]
\begin{center}
\begin{tabular}{cc}
{\bf (a)} & {\bf (b)} \\
\includegraphics[width=.3\textwidth]{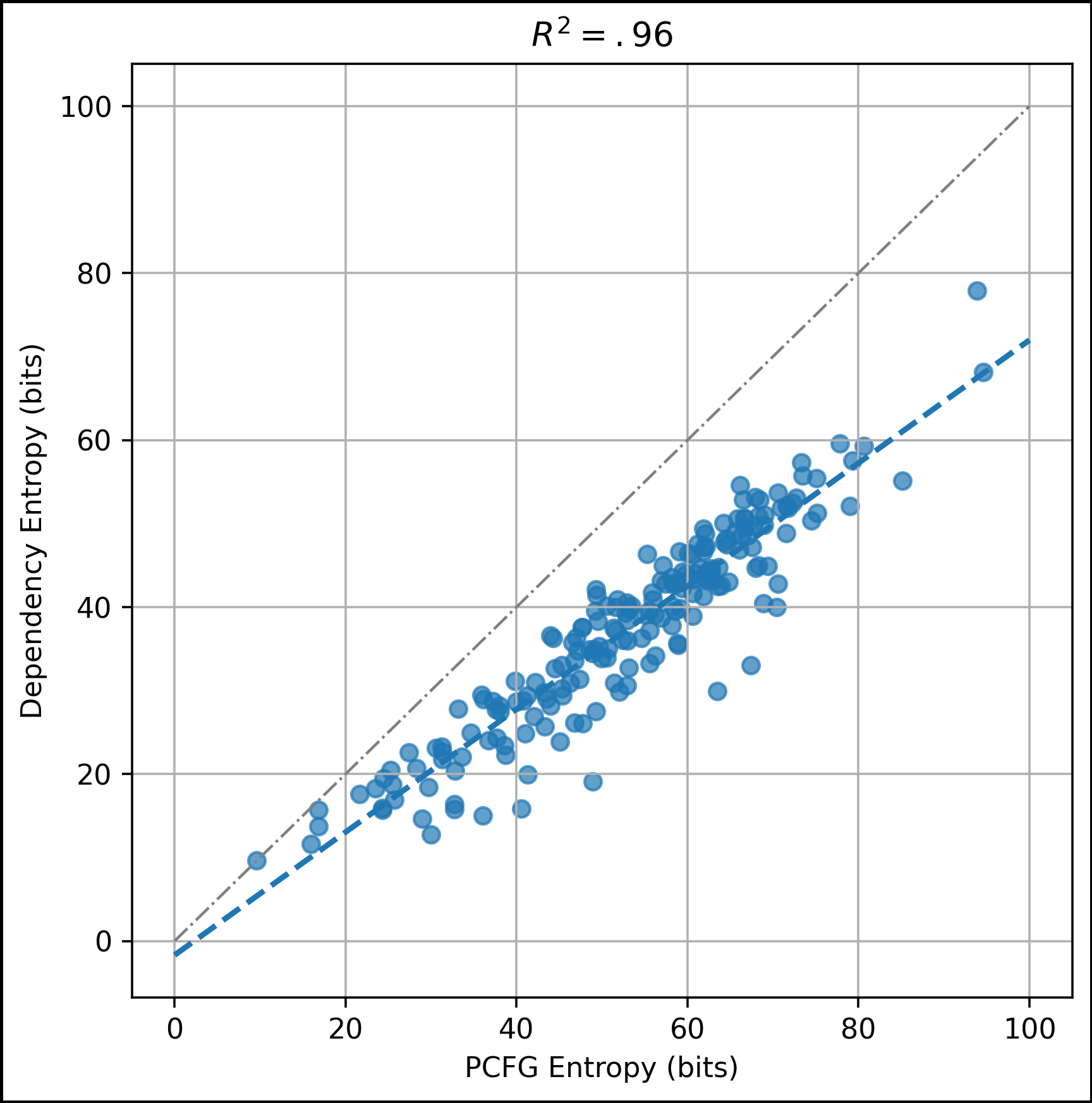} &
\includegraphics[width=.3\textwidth]{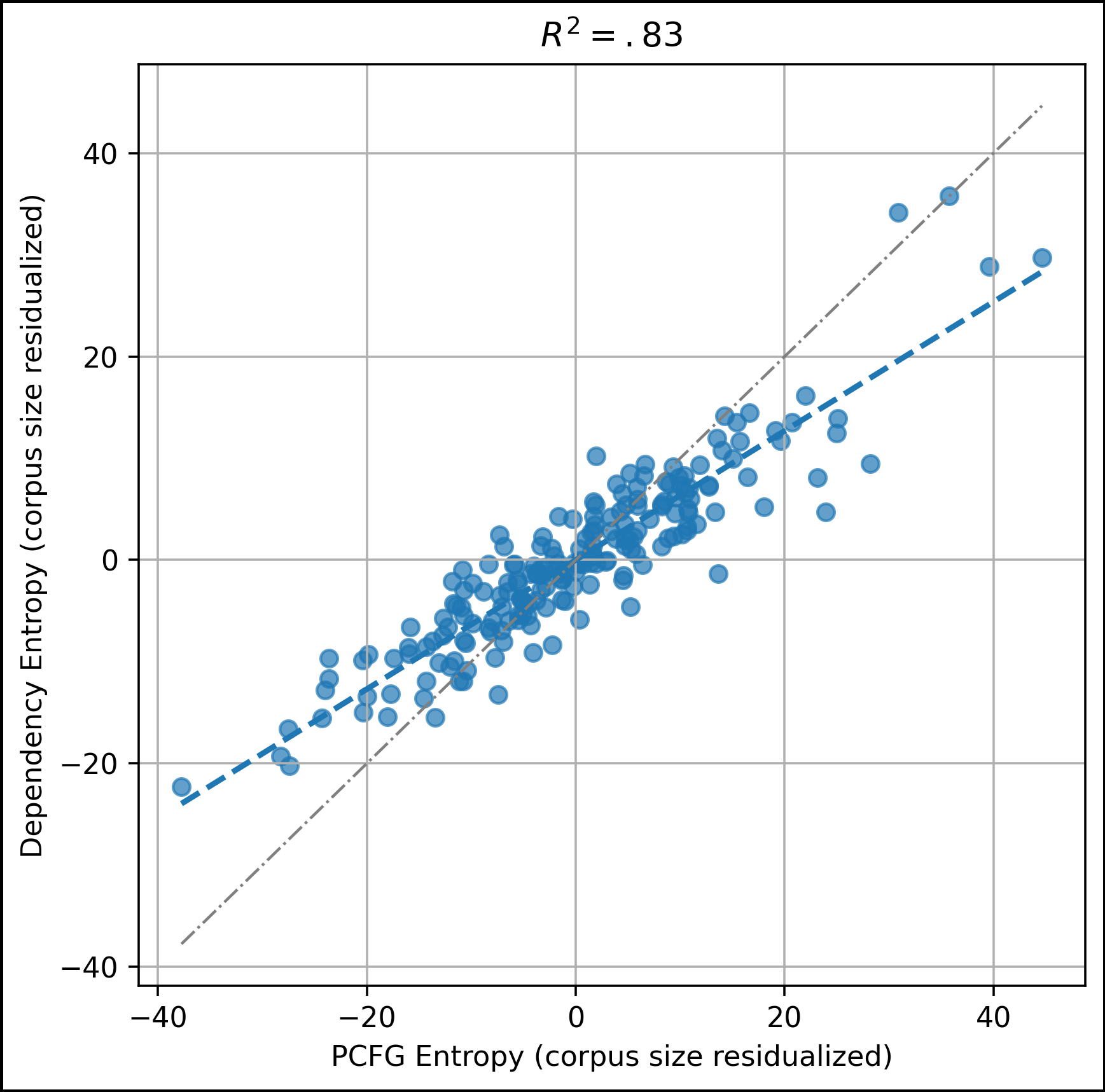}
\end{tabular}
\end{center}
\caption{Results of Corpus Analysis~\ref{sec:c1d}. Each point plots the derivational entropy estimates of a file according to the context-free (horizontal axis) and dependency (vertical axis) versions. The dashed blue lines plot linear regressions, and black dashed lines plot the identity. {\bf (a)} Relationship between the raw derivational entropies. {\bf (b)} Relationship between the derivational entropies after residualizing log corpus size.}
\label{fig:sim3}
\end{figure*}

Panel {\bf (a)} in Fig.~\ref{fig:sim3} compares the SITE entropy estimates for each corpus file according to the two grammatical theories (CFG or DG). As was expected, although the individual entropy values are different from one formalism to the other, their relative values are very strongly correlated (Pearson's $r=.98, p<.0001$). Furthermore, the line plotting the regression between both derivational entropies once again seems to go almost exactly through the origin of coordinates. This is confirmed by the lack of a significant intercept in such regression ($-1.69\pm 1.09, t[197]=-1.56, p=.12$). Refitting the regression to discard this non-significant intercept, one obtains an estimate of the relationship between the derivation entropies of $\beta \approx .71 \pm .01$ ($t[198]=130.78, p<.0001$), almost exactly what was predicted above. In other words, just knowing the derivational entropy rates in both paradigm we could account for 96\% of the variance of the derivational entropy rates of the dependency corpus using those from the context-free corpus (or vice-versa).

It could be argued that such a strong correlation could arise merely from the differences in size of the subcorpora, as the smaller subcorpora will be subject to a larger negative bias than the larger ones (i.e. in contrast with the previous section, the bias here acts in our favor). Indeed, both the context-free derivational entropy and the dependency-based one showed robust correlations with the logarithm of the corpus size (Pearson's $r=.56, p<.0001$, and $r=.69, p<.0001$, respectively). These correlations are, however, substantially weaker than that between the derivational entropies themselves. This makes it very unlikely that corpus size would account for the relationship between the two entropies. Nevertheless, to ensure that the relationship with corpus size did not confound the relations between the two entropies, I compared the corpus size residualized versions of the derivational entropies (see Analysis~\ref{sec:c1c}). Panel {\bf (b)} in Fig.~\ref{fig:sim3} plots the correlation between the residualized derivational entropies. Their correlation is a conservative estimate of the true relationship between the two entropies (i.e. by attributing to corpus size the maximum possible amount of shared variance), and it is still extremely strong (Pearson's $r=.91, p<.0001$). 

These analyses illustrate how, even if the generalization step might consider structures that would never be generated, the estimates of diversity are, in relative terms, not crucially dependent on this. The relative values of syntactic diversity are independent of whether one used CFGs or DGs to construct the treebanks. Still, one could argue that all samples used in these analyses came were extremely similar, in that they were all extracts from The Wall Street Journal, dealing with similar topics in a more or less uniform language, hence the constancy of the derivational entropy rates and its consequences. In the following analysis we will investigate whether such relations hold also for extremely heterogeneous corpora.

\subsection{A Large Heterogeneous Corpus with Unknown True Entropy}
\label{sec:c2}

Above, it was demonstrated that the SITE method is able to correctly estimate the entropy of the grammar that generated a treebank. In most real world situation, not only the true value of the entropy is unknown, but the assumption that a large treebank was generated from a single grammar is quite probably untrue. Language is dynamic, and the grammar that generated a part of a corpus possibly differs from the grammar that generated other parts of it. This is most salient in diachronic treebanks that contain samples obtained from different speakers at very different historical periods. I use the Icelandic Parsed Historical Corpus \cite*[IcePaHC v2024.03;][]{icepahc} to investigate this situation. This treebank contains a longitudinal sample of Icelandic language spanning the history of the language: from the Old West Norse of the XII century, to the contemporary Icelandic of the XXI century. The corpus contains a total of 73,051 sentences, parsed into context-free derivation trees by human annotators. The corpus is split into 61 files, each of which is a coherent text written by a single author. In addition, this corpus is also available annotated with a dependency grammar \cite*{Arnardottir:etal:2020}, as part of the Universal Dependencies project \cite[UD;][]{deMarneffe:etal:2021}. Interestingly, although the original IcePaHC and UD versions are parses of the same texts, they exhibit substantial differences in what is considered a word and what counts as a sentence. For instance, the UD version contains a total of 44,029 of sentences, about 40\% fewer than the original IcePaHC. This is the result of the UD annotation conventions requiring that several `sentences' in the original IcePaHC, are in fact taken as clauses within the same sentence in the UD annotation convention. It is interesting to investigate how the different by-design definititon of what counts as a word and what counts as a sentence would influence estimations of the derivational entropy and derivational entropy rate.

\subsubsection{Divergence of Non-stationary and Heterogeneous Samples}
\label{sec:c2a}

As mentioned above, it would be quite unrealistic to assume that all, or even a large proportion, of the texts in IcePaHC originate from a single grammar. Therefore, one should not expect to find any single meaningful value for the derivational entropy of such --likely non-existent-- grammar. To illustrate this point, I considered the the corpus in an incremental manner, adding the trees in each file of the corpus one file at a time in chronological order. At each step, the pre-terminalized trees from a file were added to the corpus, and I estimated the derivational entropy (using SITE) of the process generating the corpus by inducing a grammar from the pre-terminalized trees.

\begin{figure}[th]
\begin{center}
\includegraphics[width=.25\textwidth]{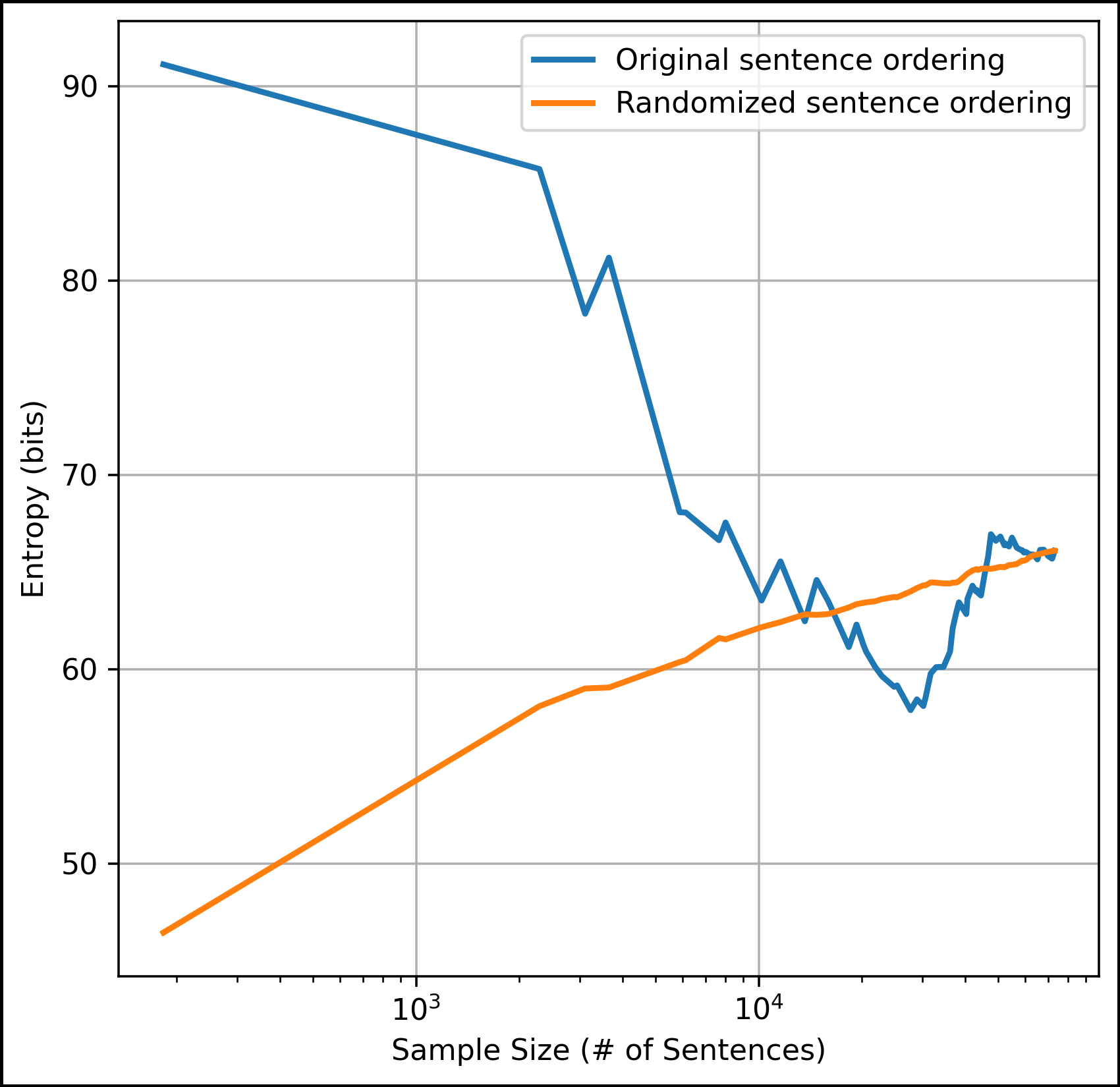} 
\end{center}
\caption{Results of Corpus Analysis~\ref{sec:c2a}. Convergence of the derivational entropies on {IcePaHC} sentence ordering across all its files. The line colors denote whether the original (blue line), or randomized sentence ordering (orange line) was used.}
\label{fig:sim4}
\end{figure}

The blue line in Figure~\ref{fig:sim4} plots the resulting entropy estimates. The first noteworthy observation is that, instead of the roughly logarithmic, monotonically increasing patterns encountered in Corpus Analyses~\ref{sec:c1a} and \ref{sec:c1b}, the estimate of derivational entropy undergoes strong oscillations along the corpus. This non-monotonic pattern indicates a lack of convergence of the derivational entropy estimate for this dataset. Before assessing convergence, however, one needs to address the issue of {\bf non-stationarity}. The analyses in sections~\ref{sec:c1a} and \ref{sec:c1b} contained sentences that had been sampled from the same grammar, and hence their statistical properties remained constant across the whole corpus. In contrast, in the current case, each file corresponds to a different author, in a different genre, written at very different historical periods. Hence, it is na\"ive to assume that the statistical properties remain in any way constant across the files. Such temporal variability of the sample's statistical properties is a textbook example of a non-stationary sequence. Among other problems, information-theoretical measures such as the entropy are meaningless in non-stationary situations.

One can force stationarity onto this corpus by randomizing the order of the sentences across all the corpus files. Such transformation has the effect of making the statistical properties of the sentences constant across the corpus, and hence truly stationary. To illustrate the effect of such manipulation, I randomized the order of all corpus sentences, and the divided the corpus into 61 chunks, each corresponding in number of sentences to one of the original corpus files. The orange line in Fig.~\ref{fig:sim4} plots the evolution of SITE estimates of derivational entropy as a function of corpus size on the randomized order corpus. Notice that, as expected, the randomization has brought back the monotonic pattern, which could lead to convergence. Note also that the end point for the randomized order estimates is 66.08 bits, identical to that obtained in the original order. The reason is that the entropy estimates themselves do not consider the order in which the sample was obtained. However, even after 73,051 sentences, SITE has not converged, this contrasts with the barely one  hundred sentences that sufficed to achieve convergence on the context-free Corpus Analysis~\ref{sec:c1a}. Furthermore, considering the logarithmic horizontal scale of the graph, convergence --if at all achievable, of which I am skeptical-- would require a corpus orders of magnitude larger, going into the hundreds of thousands or even millions of sentences. This indicates how, even with forced stationarity, the estimation methods detect the extreme level of variability stemming from the corpus heterogeneity, making it impossible to obtain a coherent entropy estimate.

\subsubsection{Convergence of Stationary Homogeneous Subsamples}
\label{sec:c2b}

In contrast with the corpus as a whole, the individual files are homogeneous, corresponding to texts generated by a single speaker, at a single historical point, and in a given genre. In this case, it is quite plausible that the parses originate from one specific grammar, whose derivational entropy could indeed be estimated. In order to assess the convergence of the derivational entropy estimates for each of the subcorpora, I randomized the order of the sentences within each file.  Each of the randomized-order files was subdivided into ten parts of equal length, and the SITE derivational entropies were computed incrementally for each of the files, adding one tenth of the file at a time.

As in the previous section, the order randomization forces stationarity within the individual files. This randomization implies discarding the discourse aspects of grammar use. Discourse makes samples of human language inherently non-stationary. For instance, if one simply looked at MLU, one would find that it tends to increase along a text. Our focus, however, is just syntax, narrowly defined in the classical sense as not extending beyond the sentence \cite[see][for views that extend syntax beyond the individual sentence]{DuBois:2014}. By forcing stationarity we are discarding any super-sentential context. This enables us to gauge the extent to which convergence for a single underlying source of syntactic rules can be achieved. Notice that trying to compute any entropy measure at a level higher than the syntax is very problematic; entropy is not even defined for non-stationary sequences.

\begin{figure}[ht]
\begin{center}
\includegraphics[width=.5\textwidth]{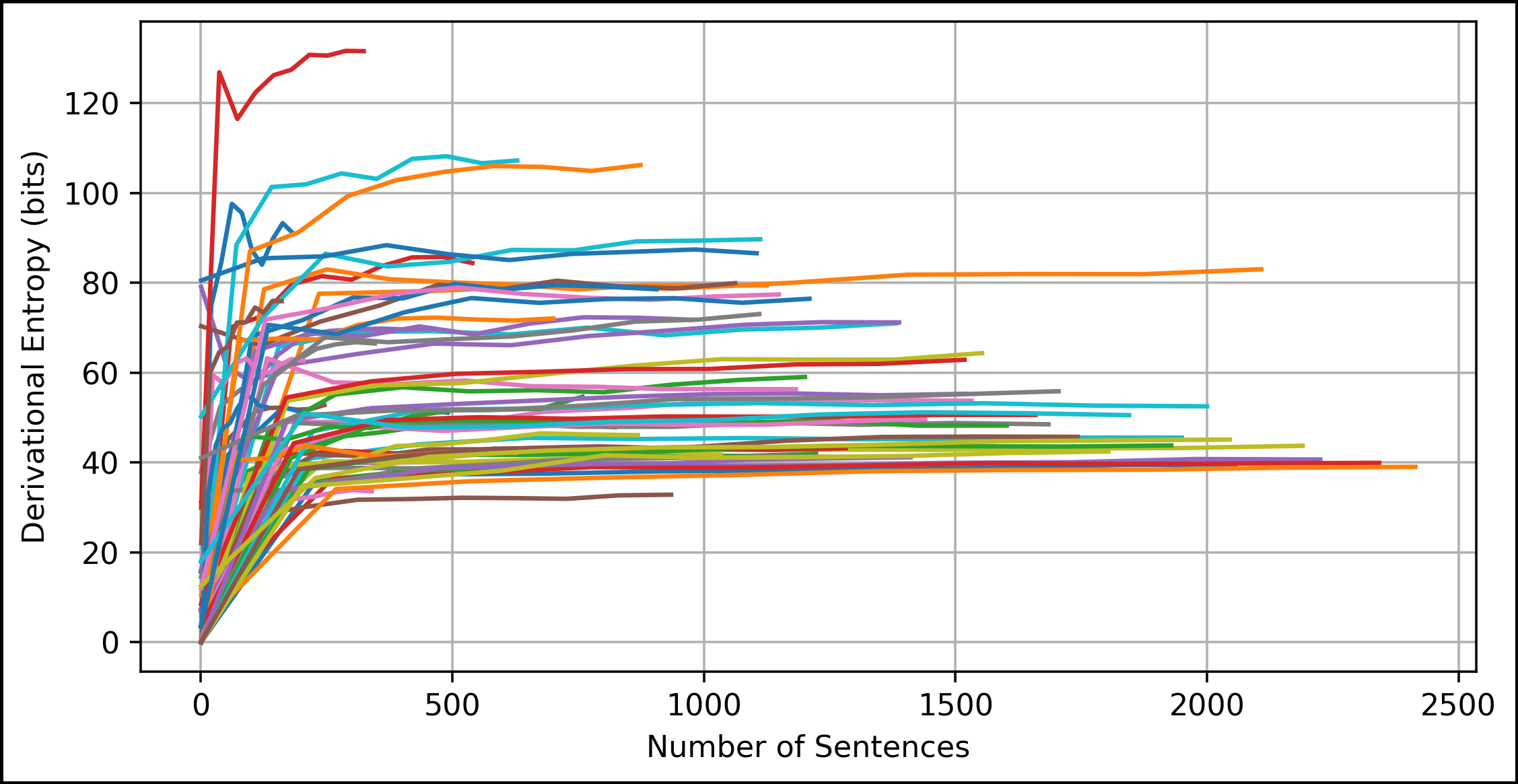} 
\end{center}
\caption{Results of Corpus Analysis \ref{sec:c2b}. Convergenge of the SITE derivational entropy estimates for each of the individual documents in {IcePaHC} (each line corresponds to a single document).}
\label{fig:sim4b}
\end{figure}

Fig.~\ref{fig:sim4b} plots the evolution of the SITE derivational entropy estimates. Except perhaps for some of the shortest files, they all seem to either have converged, or be very close to it. In contrast with the large heteregeneous corpus as a whole, the individidual files are sufficiently homogeneous to be considered as originating each from a single grammar, whose derivational entropy can indeed be estimated.

\begin{figure*}[th]
\begin{center}
\begin{tabular}{cc}
{\bf (a)} & {\bf (b)} \\
\includegraphics[width=.3\textwidth]{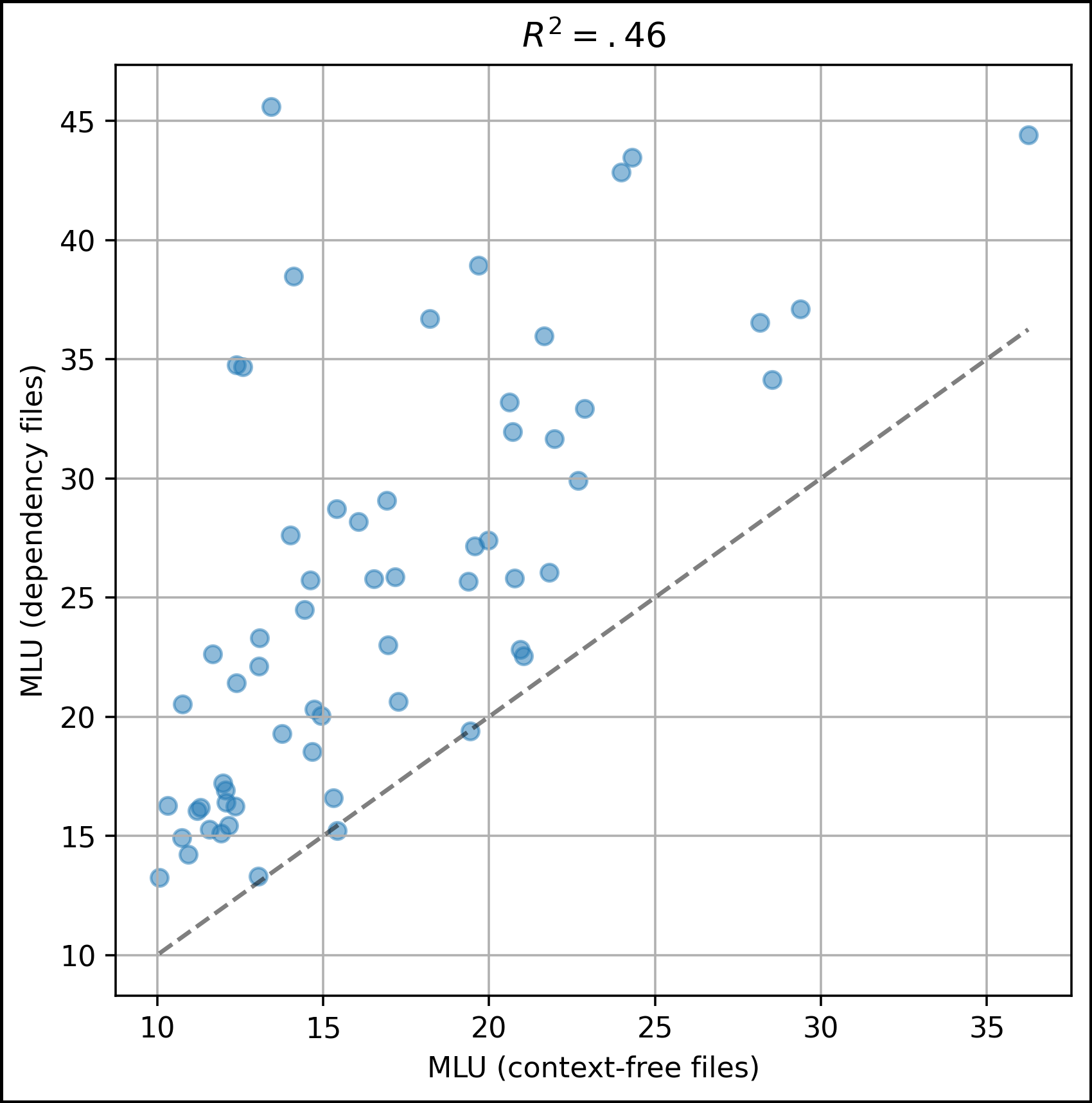} &
\includegraphics[width=.3\textwidth]{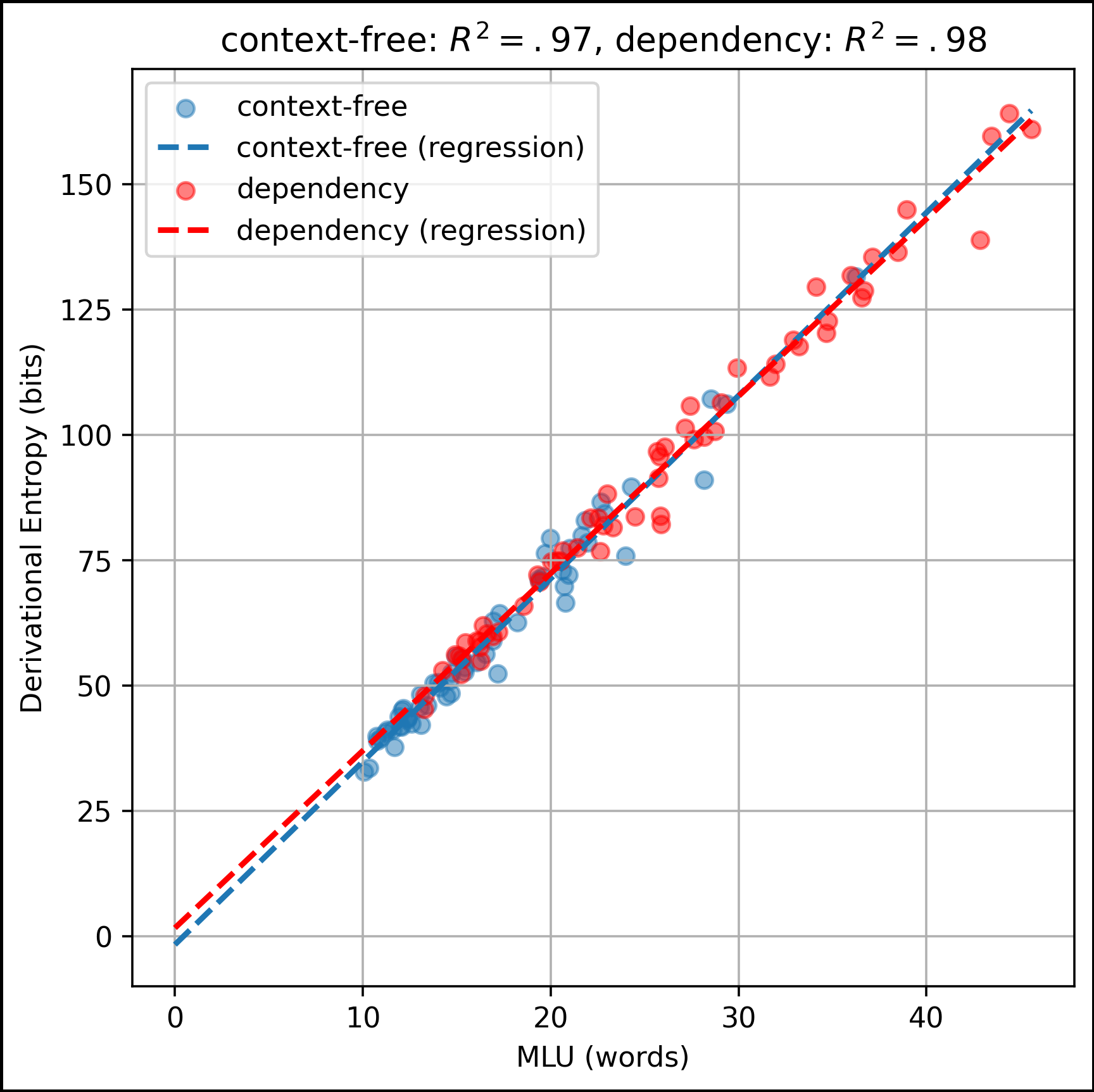}
\end{tabular}
\end{center}
\caption{Results of Corpus Analysis~\ref{sec:c2c}. {\bf (a)} Relationship between the MLUs of each of the 61 individual files of IcePaHC, comparing the CFG and UD versions of the corpus. The dashed line is the identity. {\bf (b)} Relationship between the MLU and the derivational entropies for each file in the original IcePaHC (blue dots and regression line) and the UD version of the corpus (red dots and regression line).}
\label{fig:sim4c}
\end{figure*}

\subsubsection{Constancy of Derivational Entropy Rates across Heterogeneous Corpora}
\label{sec:c2c}

As seen above, each of the individual files in IcePaHC plausibly originates from a single grammar, whose derivational entropy converges. The question is whether the apparently constant derivational entropy rate will hold as well for these very heterogeneous samples. After all, it is indeed possible that the roughly constant derivational entropy rate found in section~\ref{sec:c1c} were the result of the Wall Street Journal samples used being extremely homogeneous.

The availability of the UD version of this treebank enables further exploration the effects of heterogeneity. Although the actual texts of the files in the corpora are the same, the units of syntactic analysis --words and sentences-- are defined using different criteria in these two version. Panel {\bf (a)} in Fig.~\ref{fig:sim4c} compares the MLUs of the original IcePaHC files with those of their corresponding UD-parsed versions. As one would expect, combining multiple sentences from IcePaHC into single multi-clause sentences in the UD version results in the MLUs being substantially larger in the latter than in the former. As a result of the difference in criteria, the UD corpus has fewer sentences, and these sentences are longer. Some correlation between the MLUs across both versions does remain, but it only accounts for less than half of the MLU variance.

The derivational entropies were estimated using SITE for each of the files in each of the two corpora.  The blue dots in panel {\bf (b)} in Fig.~\ref{fig:sim4c} plot the derivational entropies as a function of the corresponding MLU for the original CFG version. The correlation is remarkably strong (Pearson's $r=.99, p<.0001$). Following the predictions, the regression (blue dashed line) hits almost exactly on the origin of coordinates, reflecting the absence of a significant intercept term ($-1.63 \pm 1.47, t[59]=-1.115, p=.27$). Excluding the intercept, the slope of this regression gives us a derivational entropy rate of $h[G] \approx 3.56 \pm .02$ bits/symbol ($t[60]=139.60, p<.0001$). Turning to the UD version of the corpus files, the correlation between their MLUs and derivational entropies (red dots in Fig.~\ref{fig:sim4b}b) was, as for the original IcePaHC, almost perfect (Pearson's $r=.99, p<.0001$). As before, the intercept term in the regression (red dashed line in the figure) was insignificant ($1.69 \pm 1.58, t[59]=1.07, p=0.29$), and the estimated derivational entropy rate was $h[G] \approx 3.59 \pm .02$ bits/symbol  ($t[60]=189.67,p<.0001$) after removing the intercept term. The degree of overlap between the estimates based on the two versions of the corpora is striking. Their derivational entropy rates are not just clearly constant within the files in each corpus, but they are also constant between the two corpora, their difference only noticeable on their second decimal digits, and their standard errors overlapping. From this, if we had two files with equivalent MLU, one from the CFG and another from the UD versions of the corpora, we can predict that their regression would have a slope of almost exactly one (i.e. $1.01$ or $.99$ depending on the directionality in which one wants to predict), indicating that their derivational entropies would be, for all practical purposes, identical.

It appears, therefore, that the derivational entropy rate stays constant within  a given annotation scheme, which is consistent with Hypothesis~2 of Section~\ref{sec:deventrate}. This is found in both of the corpora described here, as well as in the two corpora analyzed in Section~\ref{sec:c1}, and in the Switchboard I corpus studied in \citet{Moscoso:aging:2017}. However, the degree of overlap found between the two versions of IcePaHC cannot be considered a mere coincidence. In fact, the two versions of IcePaHC were not separately annotated according to different guidelines. Rather, an automatic conversion software \cite[\emph{UDConverter};][]{Arnardottir:etal:2020} was developed ad~hoc for creating the dependency version of the treebank automatically from the context-free original. That the change of grammatical paradigm was performed without any loss of information (i.e. keeping the derivational entropy rate constant), bears witness to the quality of this conversion tool \cite*{Arnardottir:etal:2023}.

\section{General Discussion}
\label{sec:discussion}

\subsection{Convergence of SITE}

I have shown that SITE provides an accurate estimation for the derivational entropy of the grammar from which a corpus has been sampled. This confirms the validity of the methods originally introduced by \citet{Moscoso:2014,Moscoso:aging:2017}, providing an explicit account of the circumstances under which the estimator can or cannot converge, as well as extending the methods to dependency treebanks. The analyses in section~\ref{sec:c1} demonstrated that, even for large, realistic grammars --either context-free or dependency-- the derivational entropy converges very fast to its true value, needing just over one hundred sentences in the context-free case, and closer to one thousand in the dependency case. In both cases, this is a substantial improvement over ML methods, which would only approach convergence with more than an order of magnitude additional sentences. This makes SITE suitable for investigating diversity in the small samples often available in many fields.

When one assumes a sample originates from a probabilistic grammar, of whichever type, one is stating that there exists a particular stable probability distribution for the syntactic regularities found in the sample. When the sample comes from heterogeneous sources, however, such assumption is not realistic. As seen in the analysis in section~\ref{sec:c2a}, in high heterogeneity situations, SITE fails to converge, even with samples of tens of thousands of sentences. The lack of convergence remains even after ensuring stationarity through sentence order randomization. In itself, the lack of convergence of SITE becomes a useful tool for assessing the homogeneity of treebank samples, prior to any additional inference one intends to make on the grammar(s) that might have generated them. Note however, that when much smaller, but homogeneous, samples from the same corpus are considered separately (section~\ref{sec:c2b}), convergence is indeed achieved quite fast.

\subsection{MLU is the Primary Measure of Syntactic Diversity \ldots}

It is not without some irony that, after using substantial theoretical apparatus for developing a sophisticated, information theoretical measure of syntactic diversity, good, old, and humble MLU \cite{Nice:1925} comes out not just unscathed, but rather reinforced on both theoretical and empirical grounds. I have demonstrated that MLU is more than just a ``proxy'' measure of syntactic diversity. Instead its value is \emph{fundamentally} linked to that of the derivational entropy. It is, therefore, an explicit measure of syntactic diversity inasmuch the derivational entropy is. The strong intercorrelation between both measures is theoretically predictable to be a linear relationship without an intercept, that is, a plain direct proportionality (section~\ref{sec:MLU}). The corpus analyses (sections~\ref{sec:c1c} and \ref{sec:c2c}) confirmed this relationship, replicating the almost perfect correlation between MLU and derivational entropy that had been previously observed \cite{Moscoso:aging:2017} in additional corpora.

In light of these findings, MLU appears to be the most suitable measure of syntactic diversity in most contexts. Sure enough, one can easily find \emph{individual} examples where an increase in sentence length does not require a more complex grammar. However, such increases are not tenable in macroscopic terms. For grammars that are actually productive those cases become mere anecdotes swamped by the strong correlation. That a substantial realistic sample (i.e. not a set of cherry picked sentences) in one language has a higher MLU than another, implies beyond doubt a higher grammatical diversity. As discussed earlier, MLU has repeatedly proven its worth in investigating multiple aspects of language, from acquisition, to aging, and disease. It is by far the simplest measure to compute, requiring very little labelling of corpora, and few --if any-- theoretical commitments. It is also unbiased, and it converges on rather small samples \cite{Casby:2011}. Although the units of measurement of MLU are in general not crucial \cite{Parker:Brorson:2005}, one might want to change the unit of measurement of MLU depending on the typological properties of the languages under study. For instance, in polysynthetic languages, where the definition of what counts as ``one word'' is less than clear, a morpheme-based or even syllable-based count would be a more natural choice \cite[see, e.g.][for Inuktitut]{Allen:Dench:2015}. Whatever the units used, \emph{MLU directly measures derivational entropy} in those units. These units map linearly into classical information theoretical units by a factor corresponding to the derivational entropy rate.

\subsection{\ldots Complemented by the Derivational Entropy Rate}

I have introduced a new measure, the derivational entropy rate, that provides a natural complement to MLU. In domains where the actual value of the derivational entropy is required explicitly, the derivational entropy rate becomes of crucial importance. As seen in the corpus analyses, and also in the results of \citet{Moscoso:aging:2017}, this measure is constant for corpora of a single language annotated using the same guidelines, supporting Hypothesis~2 of Section~\ref{sec:deventrate} (and falsifying Hypothesis~1). Therefore, for estimating the actual derivational entropy of a group of texts within a language, it suffices with estimating it using SITE for a single annotated text (ensuring its convergence). From this text, the derivational entropy ratio can be computed as a simple division. That single derivational entropy rate can then be used in conjunction with the MLUs to accurately reconstruct the derivational entropies of the remaining texts. These additional texts do not even need to be syntactically annotated, it suffices with \emph{assuming} they would be annotated \emph{using the same convention}. This approach is crucial when comparing the complexity of samples in different languages; as the correlation between length and derivational entropy is expected to vary across languages, one needs to choose (theoretically motivated) syntactic annotation schemes for each language that can be considered comparable. Then, the derivational entropies of the text can be explicitly compared across languages.

Another domain that requires the actual explicit derivational entropy rate values is for the estimation of grammar ambiguity and expected parsing difficulty \cite{Corazza:etal:2013}. The derivational entropy rate of a grammar is the rate at which the diversity of the structures of the grammar grows when adding one symbol to a sequence. Derivational entropy rate is closely related to the {\bf entropy rate} \cite{Shannon:1948} of the stochastic language generated by the grammar's closure \cite[i.e. the infinite process generated by concatenating strings randomly sampled from the grammar;][]{Soule:1974}. In PCFGs, the entropy rate is given by 
\begin{equation}
h_s[G] = \frac{H_s[G]}{MLU[G]}
\end{equation}
where $H_s[G]$ is the {\bf sentential entropy} of the grammar, that is the entropy of the distinct strings (as opposed to the distinct trees) that the grammar generates \cite[see,][for properties]{Kuich:1970}. A grammar's sentential entropy is upper bounded by the derivational entropy ($H_s[G] \leq H[G]$), with equality if and only if the grammar is unambiguous. The difference 
\begin{equation}
R[G] = H[G] - H_s[G] \geq 0
\end{equation}
is referred to as the {\bf equivocation} of the grammar; it measures the degree to which a grammar is ambiguous \cite{Soule:1974}. By parallelism, we can define the {\bf equivocation rate} of a grammar,
\begin{equation}
r[G] = h[G] - h_s[G] \geq 0
\end{equation}
which is zero if and only if the grammar is unambiguous. Estimation of $r[G]$ from a corpus is possible given an adequate estimate of the entropy rate $h_s[G]$ (which cannot be computed exactly from a grammar). Whenever the objective is to compare the parseability of multiple corpora that use the same annotation scheme, one only needs to estimate the derivational entropy rate for one of the corpora (it will be the same for all of them), and then individually estimate the entropy rates $h_s[G]$, greatly simplifying algorithms such as that proposed by \citet{Corazza:etal:2013}. Notice that, if as hypothesized above, the derivational entropy rate $h[G]$ is fully determined by a specific syntactic annotation convention, such convention also sets an upper bound for the degree of ambiguity that might be found.

The above has one important consequence: Within a given annotation convention, the only ways to reduce the syntactic ambiguity would be either using shorter sentences (decreasing the MLU), or --counterintuitively-- \emph{increasing} the sentential entropy rate ($h_s[G]$). In other words, the only way of reducing the syntactic ambiguity of sentences of a given length would be to increase the unpredictability of the sequence of symbols; unpredictable sequences of words should be syntactically less ambiguous than predictable ones. This has important implications both for natural language processing \cite[e.g.][]{Corazza:etal:2013}, and for human language comprehension and production \cite[e.g.][]{Hale:2003,Hale:2006,Linzen:Jaeger:2014,Sy:etal:2023} that are well worth studying further.

\subsection{A Conjecture: Consistent Families of Grammars and Annotation Invariance}

Above, I have found support for Hypothesis~2 of Section~\ref{sec:deventrate}. The derivational entropy rate is constant across diverse subcorpora in four different corpora: the CFG and DG versions of the Penn Treebank subsample, the original IcePaHC, and its UD version. In addition, this constancy was also found across the conversations in the Switchboard I corpus \cite{Moscoso:aging:2017}. In summary, I have repeatedly found this in all corpora I have investigated, including some additional ones not reported here for the sake brevity. 

When researchers define the  convention that will be used for syntactically annotating a treebank, they are implicitly defining a family of possible grammars (and discarding many others). For instance, in the case of Icelandic, the  guidelines that were given to the individual IcePaHC annotators defined: an alphabet (comprised of the allowed pre-terminal symbols), a set of non-terminal symbols $N$, a root symbol, and a set of \emph{possible rules}. It is not that the guidelines explicity define which individual rules are ``legal'', but they constrain which types of rules can or cannot be applied, and how specific grammatical phenomena should be annotated. Formally, we can take the rule constraints in an annotation guideline as defining a, possibly infinite, set of grammar rules, which can be taken as a skeleton Skel$[G]$ for a finite family of grammars $\mathbf{D}_f[G]$ (see section~\ref{sec:finitefamily}). Given a grammar skeleton Skel$[G]$, I define a {\bf consistent family of grammars} as
\begin{equation}
C[G] = \{\text{Dist}[G'] \, | \, \text{Dist}[G'] \in \mathbf{D}_f[G] \text{ and } h[G'] = \alpha\}
\end{equation}
where $\alpha>0$ is a fixed parameter. I will call a set of guidelines for syntactically annotating a corpus {\bf consistent} if it defines a consistent family of grammars.  I conjecture that adequate corpus annotation schemes and guidelines are consistent in this sense.

When one has two alternative consistent annotation schemes $C_1$ and $C_2$, one can compare the values of their derivational entropy rates $\alpha_1$ and $\alpha_2$. If $\alpha_1 > \alpha_2$, that is, sentences annotated using $C_1$ have, on average, higher derivational entropies than the same sentences annotated according to $C_2$. This implies that annotating a sentence using $C_2$ is an irreversible process with respect to its annotation in $C_1$, that is to say, this process loses information about possible additional parses of the same sentence. On the other hand, by its effect in reducing the equivocation rate, the syntactic ambiguity will be smaller for sentences parsed using $C_2$ than it would have been with $C_1$, and this implies that parsing using $C_2$ will be easier and more accurate than using $C_1$ \cite{Corazza:etal:2013}.

Importantly, when several linguistic samples or corpora are annotated with two different consistent annotation schemes, their derivational entropies will exhibit \emph{annotation invariance}: The specific annotation scheme used is irrelevant for the relative values of the derivational entropies, i.e. $H[G_1] = \alpha_1/\alpha_2 \, H[G_2]$. In other words, when comparing the syntactic diversities of samples within a consistent scheme, the choice of specific scheme will not affect the results. In relative terms, syntactic diversity is an inherent property of the linguistic samples, not of the specific theoretical choice of syntactic representation.

\appendix

\section{Bias-reduced Entropy Estimators}

The Coverage-Adjusted Entropy Estimator \cite[CAE;][]{Chao:Shen:2003} directly address the two sources of bias in ML entropy estimates: On the one hand, when the probabilities used in the definition of entropy are ML estimates, these are themselves positively biased. The relative frequencies of events only take into account those events that are actually attested in the sample. For an event observed $f_i$ times on a sample of size $n$, its ML probability estimate is given by $\hat{p}_i = f_i / n$, so that the sum of all $\hat{p}_i$ over the $V$ different types attested is one:
\begin{equation}
\sum_{i=1}^{V} f_i = n, \qquad \sum_{i=1}^{V} \hat{p}_i = \sum_{i=1}^{V} \frac{f_i}{n} = 1
\end{equation}
Of course, this expression distributes among the $V$ observed types all the probability mass that actually correspond to those types that are not impossible, but were not documented in the sample. It is long known that the ML probability estimates can be optimally corrected using the Good-Turing \cite[GT;][]{Good:1953} estimates, which are given by\footnote{This is a simplified version of the full Good-Turing estimator, which is found to be sufficient for entropy estimation \protect\cite{Chao:Shen:2003}.}
\begin{equation}
\tilde{p}_i = \left(1-\frac{f_1}{n}\right) \hat{p}_i \label{eq:goodturing}
\end{equation}
where $f_1$ denotes the number of types that were observed exactly once in the sample (i.e. the number \emph{hapax legomena}). The GT estimates reduce the value of the ML estimates, in turn increasing the value of the ML entropy estimate, but sill underestimating the true entropy values. If $H$ denotes the true entropy value, and $\mathbb{E}\big[H_{\mathrm{ML}}\big]$ and $\mathbb{E}\big[H_{\mathrm{GT}}\big]$ 
are, respectively, the expected values of the ML and GT estimates (which are obtained by plugging the corresponding probability estimate into the classical entropy equation),
\begin{equation}
\mathbb{E}\big[H_{\mathrm{ML}}\big] \leq \mathbb{E}\big[H_{\mathrm{GT}}\big] \leq H
\end{equation}
although both estimators are consistent, converging to the true $H$ value with probability 1 for infinite sample sizes, they are both negatively biased.

The other source of bias, shared by the ML and CAE estimators, are the missing terms (corresponding to unseen, but possible, events) in the classical entropy equation. CAE also accounts for this by, on top of using GT-corrected probability estimates, changing the entropy equation itself,
\begin{equation}
H_{\mathrm{CAE}} = -\sum_{i=1}^{k} \frac{\tilde{p}_i \log(\tilde{p}_i)}{1-(1-\tilde{p}_i)^n}
\label{eq:chao2003}
\end{equation}
Notice that the numerators of this expression are the usual terms in the entropy equation. The denominators are an additional correction to account for the missing terms. Using this equation has been shown to produce entropy estimates that are less biased and faster converging that both ML and GT estimates (and in fact many other methods \cite[see,][]{Vu:Yu:Kass:2007}).

Chao and her colleagues have more recently proposed a new improved estimator \cite[CWJ;][]{Chao:etal:2013}. This estimator exploits properties of the accumulation curve of the number of species observed in an ecosystem (i.e. the species accumulation curve). It is given by
\begin{equation}
H_{\mathrm{CWJ}}  = \sum_{1 \leq F_i \leq n-1} \Bigg[ \frac{F_i}{n}  \Bigg( \sum_{k=F_i}^{n-1}\frac{1}{k} \Bigg) \Bigg] -
\frac{f_1}{n}(1-A)^{1-n}\Bigg[\log(A)+\sum_{r=1}^{n-1}\frac{1}{r}(1-A)^r \Bigg]
\label{eq:chao2013}
\end{equation}
where $F_i$ are the frequencies observed, and
\begin{equation}
A =  \left\{
\begin{array}{ll}
\frac{2f_2}{(n-1)f_1+2f_2} & \text{if } f_2 > 0\\
\frac{2}{(n-1)(f_1-1)+2} & \text{if } f_2=0, f_1>0\\
1 & \text{if } f_1 = f_2 = 0
\end{array}\right.
\end{equation}
with $f_1$ and $f_2$ being the number of types that where encountered exactly once or twice respectively (i.e. the numbers of \emph{hapax legomena} and \emph{dis legomena}). 

\section*{Acknowledgements}
The author is indebted to Prof. John W. DuBois, Dr. Enrique Amig\'o Cabrera, Suchir Salham, and Paul Siewert for proofreading earlier versions of this article, comments and discussion of the ideas, and to Dr. Hinrik Hafsteinsson for resolving doubts on the differences in sentence segmentation between the original and UD versions of IcePaHC.

\bibliographystyle{compling}
\bibliography{pcfg-entropy}

\end{document}